%% file: main.tex
\documentclass[sigconf]{acmart}

\AtBeginDocument{%
  }
\usepackage{anyfontsize}
\usepackage{newfloat}
\usepackage{listings}
\usepackage{epsfig}
\usepackage{amsmath}
\usepackage{bm}
\usepackage{algorithm}
\usepackage{algorithmic}
\usepackage{siunitx}
\usepackage{fontawesome}
\usepackage{bbding}
\usepackage{pifont}
\usepackage{wasysym}
\usepackage{amsfonts}
\usepackage{mathtools}
\usepackage{makecell}
\usepackage{enumitem}
\usepackage{caption}
\usepackage{tcolorbox}
\usepackage{multirow}
\usepackage{colortbl}
\usepackage{subcaption}

\def\dsname{\textit{RealSyn}}
\definecolor{blue_ours}{HTML}{D8ECD1} 
\definecolor{green_ours}{rgb}{0.53, 0.83, 0.64}
\definecolor{dt}{gray}{0.7}
\definecolor{lightgray}{gray}{0.91}
\definecolor{cvprblue}{rgb}{0.21,0.49,0.74}

\newcolumntype{x}[1]{>{\centering\arraybackslash}p{#1pt}}
\newcolumntype{y}[1]{>{\raggedright\arraybackslash}p{#1pt}}
\newcolumntype{z}[1]{>{\raggedleft\arraybackslash}p{#1pt}}
\newcommand{\app}{\raise.17ex\hbox{$\scriptstyle\sim$}}

\definecolor{baselinecolor}{gray}{.9}

\newcolumntype{=}{>{\global\let\currentrowstyle\relax}}
\newcolumntype{^}{>{\currentrowstyle}}

\definecolor{dt}{gray}{0.2}  %

\definecolor{lightgray}{gray}{0.91}
\newcommand*{\belowrulesepcolor}[1]{%
  \noalign{%
    \kern-\belowrulesep 
    \begingroup 
      \color{#1}%
      \hrule height\belowrulesep 
    \endgroup 
  }%
} 
\newcommand*{\aboverulesepcolor}[1]{%
  \noalign{%
    \begingroup 
      \color{#1}%
      \hrule height\aboverulesep 
    \endgroup 
    \kern-\aboverulesep 
  }%
}

\copyrightyear{2025}
\acmYear{2025}
\setcopyright{cc}
\setcctype{by}
\acmConference[MM '25]{Proceedings of the 33rd ACM International Conference on Multimedia}{October 27--31, 2025}{Dublin, Ireland}
\acmBooktitle{Proceedings of the 33rd ACM International Conference on Multimedia (MM '25), October 27--31, 2025, Dublin, Ireland}\acmDOI{10.1145/3746027.3755098}
\acmISBN{979-8-4007-2035-2/2025/10}
\begin{document}

\title{\includegraphics[width=0.5cm, height=0.7cm]{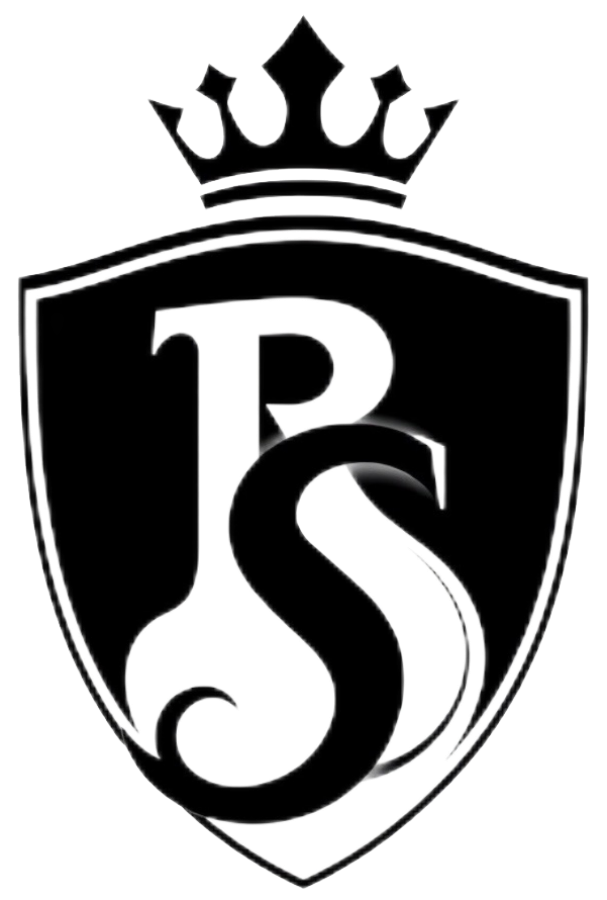} \textit{RealSyn}: An Effective and Scalable Multimodal Interleaved Document Transformation Paradigm}

\author{Tiancheng Gu$^\text{\ding{170}}$\textsuperscript{*}, Kaicheng Yang$^{\text{\ding{171}}}$\textsuperscript{*}, Chaoyi Zhang$^\text{\ding{170}}$, Yin Xie$^{\text{\ding{171}}}$, Xiang An$^{\text{\ding{171}}}$, \\ Ziyong Feng$^{\text{\ding{171}}}$, Dongnan Liu$^{\text{\ding{170}}}$, Weidong Cai$^{\text{\ding{170}}}$\textsuperscript{$\dagger$}, Jiankang Deng$^{\text{\ding{105}}}$\textsuperscript{$\dagger$}\\
$^{\text{\ding{170}}}$The University of Sydney $^{\text{\ding{171}}}$DeepGlint $^{\text{\ding{105}}}$Imperial College London \\
\texttt\tiny{tigu8498@uni.sydney.edu.au, kaichengyang@deepglint.com} \\
{\faGlobe\ \href{https://garygutc.github.io/RealSyn}{\textcolor{magenta}{\textit{Project Page}}}}  \hspace{2cm}
{\faGithub\ \href{https://github.com/deepglint/RealSyn}{\textcolor{magenta}{\textit{Code}}}}
}
\thanks{\textsuperscript{*} Equal Contribution \\
\textsuperscript{$\dagger$} Corresponding Author}

\renewcommand{\shortauthors}{Tiancheng Gu, Kaicheng Yang et al.}

\input{sections/0Abstract}



\maketitle

\input{sections/1Introduction}

\input{sections/2Related_works}
\input{sections/3Methods}

\input{sections/4Experiments}
\input{sections/5Ablation_study}

\input{sections/6Conclusion}


\bibliographystyle{ACM-Reference-Format}
\balance
\bibliography{sample-base}

\clearpage
\input{sections/Xappendix}

\end{document}

%% file: sections/0Abstract.tex
\begin{abstract}
\sloppy
After pre-training on extensive image-text pairs, Contrastive Language-Image Pre-training~(CLIP) demonstrates promising performance on a wide variety of benchmarks. However, a substantial volume of multimodal interleaved documents remains underutilized for contrastive vision-language representation learning. To fully leverage these unpaired documents, we initially establish a Real-World Data Extraction pipeline to extract high-quality images and texts. Then we design a hierarchical retrieval method to efficiently associate each image with multiple semantically relevant realistic texts. To further enhance fine-grained visual information, we propose an image semantic augmented generation module for synthetic text production. Furthermore, we employ a semantic balance sampling strategy to improve dataset diversity, enabling better learning of long-tail concepts. Based on these innovations, we construct \textit{RealSyn}, a dataset combining realistic and synthetic texts, available in three scales: 15M, 30M, and 100M. We compare our dataset with other widely used datasets of equivalent scale for CLIP training. Models pre-trained on \textit{RealSyn} consistently achieve state-of-the-art performance across various downstream tasks, including linear probe, zero-shot transfer, zero-shot robustness, and zero-shot retrieval. Furthermore, extensive experiments confirm that \textit{RealSyn} significantly enhances contrastive vision-language representation learning and demonstrates robust scalability.

\end{abstract}

%% file: sections/1Introduction.tex
\section{Introduction}
\label{sec:intro}
The rapid proliferation of mobile networks and social platforms has led to exponential growth in large-scale data, offering a robust foundation for contrastive vision-language representation learning~\cite{guo2019deep,baltruvsaitis2018multimodal, guo2024regiongpt, zhu2024llms, yao2024visual, zheng2024picture, guo2024open}. By predicting the correspondence between images and description texts, Contrastive Language-Image Pre-training (CLIP)~\cite{CLIP} pre-trains separate uni-modal encoders and achieves excellent performance across a range of downstream tasks, including image captioning~\cite{li2023blip,mokady2021clipcap}, object detection~\cite{wu2023cora,lin2023gridclip}, and semantic segmentation~\cite{he2023clip,zhou2023zegclip,lin2023clip}.
Notably, the representational capacity of such models improves significantly with the scale of the training dataset~\cite{li2024scaling}. Designing a new data paradigm to expand the scale and diversity of image-text datasets, thereby continuously improving the capabilities of pre-trained vision-language representations, represents a significant and challenging issue.

\begin{figure}[t!]
    \centering
    \includegraphics[width=0.95\linewidth]{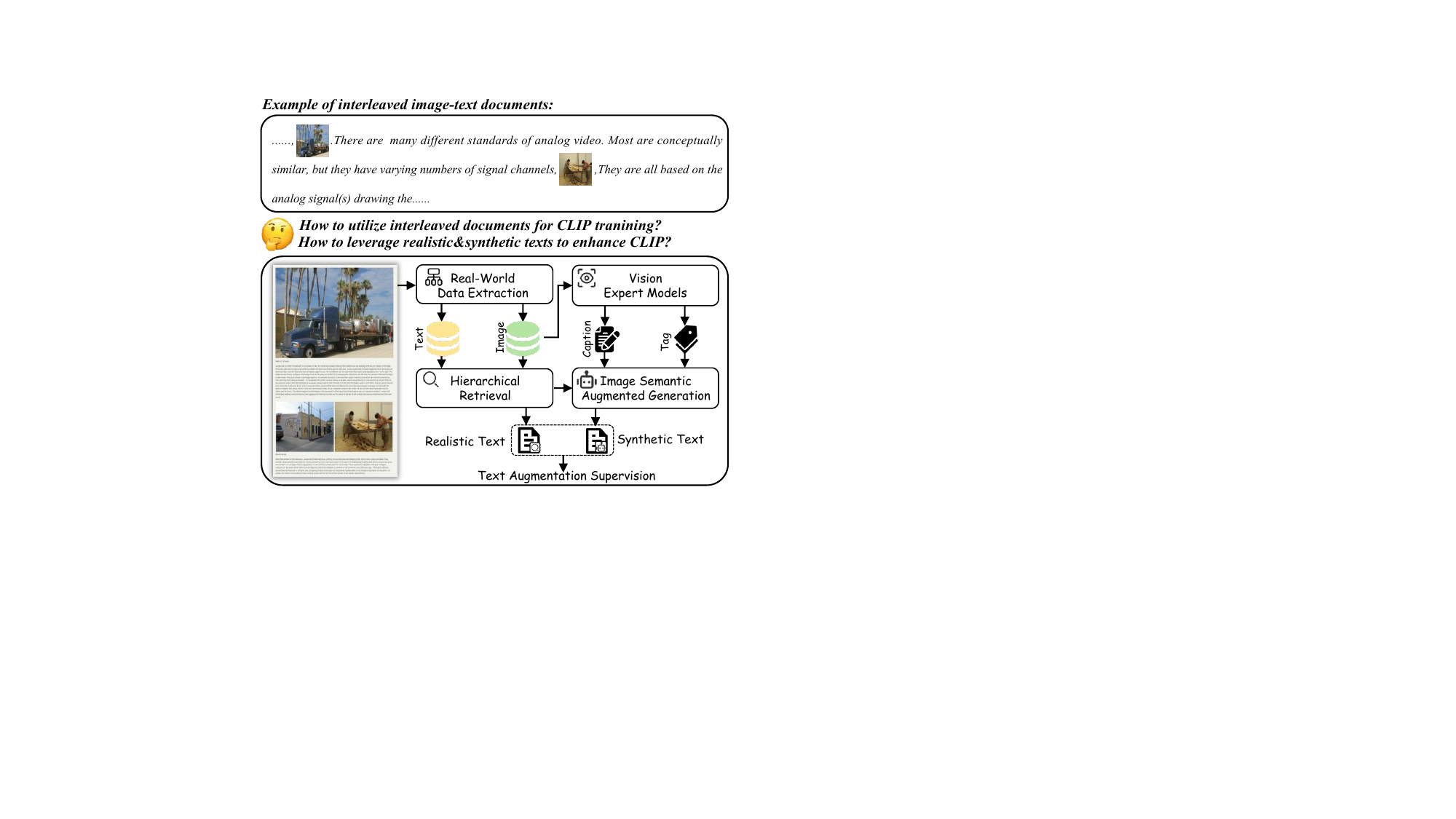}
    \caption{Multimodal interleaved documents are unsuitable for CLIP training. We construct distinct image-text pairs from such documents via retrieval and generation.}
    \label{fig: motivation}
    \vspace{-5mm}
\end{figure}

\begin{figure*}[t!]
    \centering
    \includegraphics[width=\linewidth]{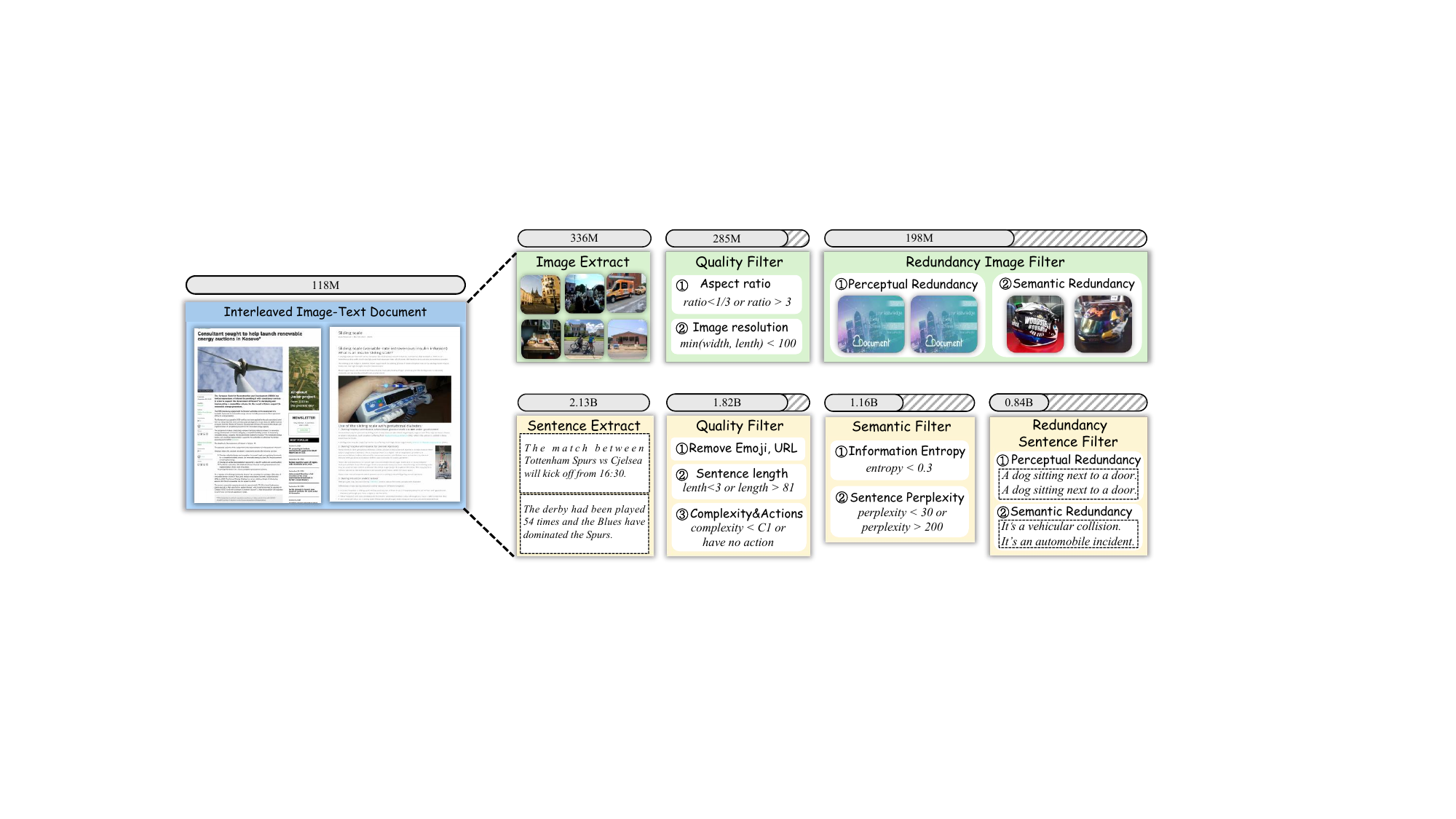}
    \vspace{-3mm}
    \caption{The Real-World Data Extraction pipeline to extract high-quality images and texts from interleaved image-text documents.}
    \vspace{-3mm}
    \label{fig: data_filtering}
\end{figure*}

In recent years, the introduction of large-scale image-text pair datasets~\cite{laion400M, laion5B, YFCC100M, kakaobrain2022coyo700m, Datacomp} has substantially advanced vision-language representation learning. Due to download failures and non-English captions in the YFCC~\cite{YFCC100M} dataset, DeCLIP~\cite{DECLIP} reprocesses a new version of the YFCC15M dataset. LAION400M~\cite{laion400M} provides 400 million image-text pairs collected from the web and filtered using CLIP similarity. COYO700M~\cite{kakaobrain2022coyo700m} offers 747 million high-quality pairs curated through advanced filtering strategies, including CLIP similarity, watermark detection, and aesthetic scoring. Despite these advancements, a substantial amount of non-paired data, particularly interleaved image-text documents~\cite{MMC4, Obelics, omnicorpus} containing multiple images and text paragraphs without explicit correspondence, remains incompatible with conventional vision-language representation learning methods.

Recent studies~\cite{laclip, yu2024capsfusion, DreamLIP, Veclip} aim to enhance the quality of open-source image-text datasets using existing models. For instance, LaCLIP~\cite{laclip} leverages large language models (LLMs) to rewrite raw captions for better alignment with images. CapsFusion~\cite{yu2024capsfusion} fine-tunes an LLM using a ChatGPT-generated instruction dataset to mitigate hallucination issues. DreamLIP~\cite{DreamLIP} re-captions 30 million images with detailed descriptions using a pre-trained large multimodal model. However, these methods primarily enhance data quality by focusing on synthetic data, thereby overlooking the significance of real-world data. Moreover, the diversity and distribution of synthetic captions produced by these approaches are inherently restricted by the limitations of the underlying generative models.

In this paper, we explore two fundamental questions: \textbf{1) How to utilize multimodal interleaved documents for CLIP training. 2) How to effectively leverage both realistic and synthetic texts to enhance CLIP performance.} As depicted in Figure~\ref{fig: motivation}, we initially develop a Real-World Data Extraction pipeline to extract high-quality images and texts from real-world multimodal interleaved documents. Subsequently, we design a hierarchical retrieval method to efficiently associate each image with multiple semantically relevant texts. To improve fine-grained image understanding, we introduce a visual semantic augmented generation module for synthetic text production. Additionally, we implement a semantic balance sampling strategy to enhance dataset diversity and facilitate the learning of long-tail concepts. Leveraging these innovations, we construct the \textit{RealSyn} dataset, which incorporates both realistic and synthetic texts in three sizes: 15M, 30M, and 100M. We evaluate the \textit{RealSyn} dataset against other widely used datasets of similar scale for CLIP training. Models pre-trained on \textit{RealSyn} consistently demonstrate state-of-the-art performance on various downstream tasks, including linear probe, zero-shot transfer, zero-shot robustness, and zero-shot retrieval. Extensive experiments validate that \textit{RealSyn} is effective for contrastive vision-language representation learning and shows excellent scalability. The main contributions of this paper are summarized as follows:
\begin{itemize}[leftmargin=*]
\item We propose an \textbf{effective and scalable multimodal interleaved document transformation paradigm} for contrastive vision-language representation learning.
\item We release \textit{RealSyn}, \textbf{a large-scale semantic balanced dataset} that integrates both realistic and synthetic texts and is available in three sizes: 15M, 30M, and 100M.
\item We conduct \textbf{extensive experiments} and demonstrate the effectiveness and scalability of our proposed \textit{RealSyn} dataset for CLIP training.
\end{itemize}

%% file: sections/2Related_works.tex
\section{Related Work}
\label{sec:formatting}

\noindent{\bf Large-Scale Pre-training Dataset.} In recent years, several large-scale image-text datasets~\cite{MSCOCO, clark2017simple, goyal2017making, kakaobrain2022coyo700m, Densefusion1M} collected from the Internet have been released. The YFCC100M~\cite{YFCC100M} dataset provides a comprehensive overview of the evolution of photo and video documentation and sharing from the inception of Flickr in 2004 until early 2014. Due to download failures and non-English captions, DeCLIP~\cite{DECLIP} reprocesses a new version of the YFCC15M dataset. Additionally, the LAION400M~\cite{laion400M} dataset contains 400 million image-text pairs collected from Common Crawl and widely used in vision-language pre-training. Recent advancements have also introduced several large-scale interleaved image-text document datasets~\cite{omnicorpus, MMC4, Obelics}. The OBELICS~\cite{Obelics} dataset uses a comprehensive filtering strategy and includes 141 million web pages, 353 million associated images, and 115 billion text tokens extracted from Common Crawl. However, due to data format constraints and training inefficiencies, interleaved image-text documents are currently unsuitable for CLIP training.

\noindent{\bf Vision Language Pre-training.}
As a pioneering work in visual language pre-training, CLIP has attracted extensive attention due to its powerful zero-shot recognition and exceptional transfer learning performance~\cite{wang2024learn, tang2024amu, shao2024deil, martin2024transductive, hu2025decoupled, gu2024rwkv}. Inspired by CLIP, numerous visual-language pre-training works have been published in recent years~\cite{mu2022slip, DECLIP, ALIP}. SLIP~\cite{mu2022slip} enhances performance by combining self-supervised learning with CLIP pre-training. DeCLIP~\cite{DECLIP} increases pre-training efficiency by integrating multi-view supervision across modalities and nearest-neighbor supervision from similar pairs. To mitigate the influence of noisy data, ALIP~\cite{ALIP} introduces a gating mechanism that dynamically allocates weights to samples. Despite their advancements, these methods primarily depend on large-scale image-text pairs derived from the Internet. Recent studies~\cite{li2024scaling,wang2025scaling} demonstrate that the capabilities of CLIP enhance with the expansion of high-quality image-text datasets. Given the extensive use of internet-derived image-text data in existing datasets, there is a pressing need to develop a new data construction paradigm to further expand the scale of high-quality image-text data.

\noindent{\bf Synthetic Captions.} 
Recent works~\cite{ALIP, yu2024capsfusion, sharegpt4v} indicate that image-text pairs obtained from websites contain intrinsic noise, which directly impacts the effectiveness of vision-language pre-training. To enhance the quality of existing datasets, LaCLIP~\cite{laclip} uses the in-context learning capability of large language models to rewrite text descriptions associated with each image. CapsFusion~\cite{yu2024capsfusion} employs large language models to refine information from web-based image-text pairs and synthetic captions, improving the quality of multimodal pre-training data. 
Similarly, DreamLIP~\cite{DreamLIP} generates detailed descriptions for 30 million images using a pretrained large multimodal model. Nevertheless, these methods predominantly focus on synthetic data enhancement, neglecting the importance of real-world data. Furthermore, the diversity and distribution of synthetic captions generated by these methods are intrinsically constrained by the capabilities of the generative models employed.

%% file: sections/3Methods.tex
\section{\textit{RealSyn} Dataset}

\subsection{Real-World Data Extraction}
To transform interleaved image-text documents for vision-language representation learning, we establish a Real-World Data Extraction pipeline~(Figure~\ref{fig: data_filtering}) to extract high-quality images and texts. This pipeline consists of three steps: Data Extraction, Image Filtration, and Sentence Filtration.

\noindent{\bf Data Extraction.} We employ 118M interleaved image-text documents from the OBELICS~\cite{Obelics} as the primary data source. All images are extracted and stored in a dedicated image database, while sentences are segmented using the Natural Language Toolkit (NLTK)~\cite{nltk} and stored in a separate sentence database. This process yields 336M images and 2.13B sentences from the interleaved documents.

\noindent{\bf Image Filtration.}
After extracting 336M images, we apply a two-stage filtering process to ensure data quality and reduce redundancy. First, we discard images that meet any of the following criteria: 1) the shorter dimension is fewer than 100 pixels, or 2) the aspect ratio exceeds 3 or is below 1/3. This step removes 51M low-quality images. Next, following CLIP-CID~\cite{CLIP_CID}, we use the EVA02-CLIP E/14-plus model~\cite{eva_clip} to extract image embeddings and apply the Union-Find algorithm~\cite{union_find} to eliminate perceptually and semantically redundant images. This step removes an additional 87M images, resulting in a refined dataset of 198M high-quality images.

\noindent{\bf Sentence Filtration.}
After extracting 2.13B sentences from interleaved image-text documents, we conduct rigorous filtering based on quality, semantics, and redundancy. Initially, we eliminate sentences based on the following criteria: 1) presence of emojis or URLs; 2) sentences containing fewer than 3 or more than 81 words; and 3) following CAT~\cite{CAT_rule}, we retain samples with at least C1 caption complexity and incorporating an action. This phase reduces the corpus size from 2.13B to 1.82B sentences. Then we apply semantic filtering to the remaining sentences, eliminating those with minimal information assessed through information entropy:
\begin{equation}
\label{equation:IE}
\theta(x) = -\sum_{i=1}^n p(x_{i}) \log p(x_{i}),
\end{equation}
where $n$ denotes the number of words in a sentence, $x_{i}$ represents the $i$-th words in the sentences $x$, and $p(x_{i})$ is the probability of the word $x_i$ in the entire corpus. Based on human cognition principles and empirical experience, we filter out sentences with a score below 0.3. To further refine the corpus by removing difficult or ambiguous sentences, we use GTP2-large~\cite{gpt2} to calculate the perplexity score $\mathcal{PPL}$ for each sentence:
\begin{equation}
\label{equation:perplexity}
\mathcal{PPL}(x) = \exp \left\{-\frac{1}{t} \sum_{i=1}^{t} \log p_{\theta}(x_i \mid x_{<i}) \right\},
\end{equation}
where $t$ represents the token number of the sentence, and $p_{\theta}(x_i \mid x_{<i})$ is the likelihood of the $i$-th token given the previous tokens. We engage three human experts to determine the minimum and maximum values of the perplexity interval for high-quality sentences by comparing sentences of varying perplexity levels. Sentences within the average minimum~(30) and maximum perplexity~(200) intervals are retained. The overall semantics filtering reduces the corpus to 1.16B sentences. In the final stage, similar to redundancy image filtering, we perform both perceptual and semantic deduplication of sentences. This process results in a refined corpus of 0.84B sentences that include extensive real-world knowledge.

\begin{figure}[t]
    \centering  \includegraphics[width=\linewidth]{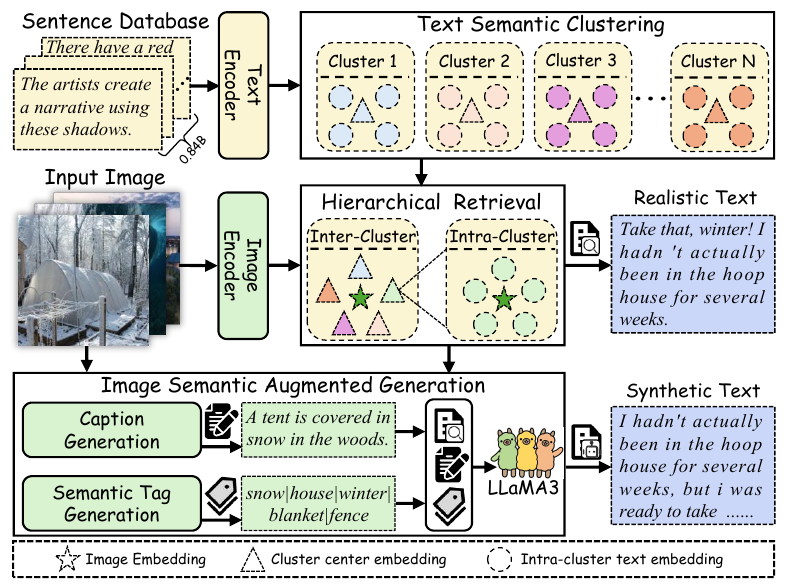}
    \vspace{-3mm}
    \caption{The architecture of our proposed framework, which constructs distinct image-text pairs from real-world data extracted from interleaved documents via retrieval and generation.}
    \label{fig: pipeline}
    \vspace{-5mm}
\end{figure}

\input{Tables/Linear_probe}
\input{Tables/Zeroshot_classification} 

\subsection{\bf Retrieval and Generation Framework} 
After extracting high-quality images and sentences from documents, we propose an efficient and scalable framework to retrieve multiple semantically relevant texts for each image and leverage large language models to integrate retrieved realistic text with fine-grained visual information and generate synthetic text. As shown in Figure~\ref{fig: pipeline}, the architecture of our framework primarily consists of three components: Text Semantic Clustering, Hierarchical Retrieval, and Image Semantic Augmented Generation.

\noindent{\bf Text Semantic Clustering.} To efficiently retrieve multiple semantically relevant texts for each image, we initially encode all sentences using the EVA02-CLIP E/14-plus~\cite{eva_clip} model. Inspired by Unicom~\cite{unicom}, we utilize the standard $K$-Means algorithm~\cite{kmeans} offline cluster all sentences into a specified number of clusters. To minimize intra-cluster and inter-cluster conflicts, we initially establish the optimal number of cluster centroids through small-scale experiments. Subsequently, we divide the 0.84 billion texts into two million clusters, utilizing efficient feature quantization techniques~\cite{johnson2019billion}.

\noindent{\bf Hierarchical Retrieval.} Given the prohibitive computational cost of direct semantic text retrieval across 0.84B sentences (requiring over 10,000 GPU-hours on 8×A100), we propose a hierarchical retrieval framework to enhance efficiency. Given an image, we perform inter-cluster retrieval to identify the most relevant cluster center for the image. Subsequently, we conduct intra-cluster retrieval within the identified cluster to obtain multiple real-world sentences semantically matched to the image. This approach achieves scalable retrieval of 198M images and 0.84B sentences within 40 GPU-hours on an 8×A100.

\noindent{\bf Image Semantic Augmented Generation.} Although the retrieved realistic texts achieve satisfactory performance, they exhibit limitations in capturing fine-grained visual semantics. To address this issue, we introduce the Image Semantic Augmented Generation module. This module initially employs the OFA~\cite{OFA} model to generate a concise caption for each image. We then integrate the open-set image tagging model RAM++~\cite{RAM_plus_plus}, which extracts object detection tags. Considering that RAM++ supports only 4,000 tags, we expand this set to 8,000 by incorporating an additional 4,000 tags derived from real-world sentences. Following CapsFusion~\cite{yu2024capsfusion}, we utilize ChatGPT4 Turbo to merge the retrieved realistic texts with concise captions and image tags to construct a 100K instruction dataset (the prompt we used is presented in the supplementary material). Subsequently, we conduct instruction tuning of the LLaMA3-8B model~\cite{llama3} using the LLaMA Factory~\cite{llamafactory} and deploy the vLLM~\cite{vLLM} for large-scale inference. Ultimately, we convert data from 118M multimodal interleaved documents into 198M image-text pairs, where each image is associated with multiple retrieved realistic texts and synthetic texts.

\input{Tables/Zeroshot_retrieval}

\input{Tables/Zeroshot_robustness}

\subsection{Semantic Balance Sampling}
To further improve the quality and diversity of our dataset, we implement semantic balancing sampling across the 198M image-text pairs. Specifically, we use EVA02-CLIP E/14-plus~\cite{eva_clip} to encode and calculate the cosine similarity between images and synthetic texts. To reduce the impact of OCR-related or mismatched pairs during pre-training, we filter out 29.7M pairs with cosine similarities scores either above 0.61 or below 0.51, as determined through human verification. Inspired by MetaCLIP~\cite{xu2023metaclip}, we introduce an easy but efficient cluster-based semantic balance sampling strategy. We cluster the image embeddings from the remaining 168.3M pairs into 1M centers. To enhance the semantic diversity of our dataset, we randomly select 20, 35, and 180 samples from clusters exceeding these thresholds, while retaining all samples from smaller clusters.  This approach culminates in the construction of the \dsname15M, \dsname30M, and \dsname100M datasets.

%% file: Tables/Linear_probe.tex
\begin{table*}[t]
    \centering
    \caption{Linear probe on 20 downstream datasets. Pre-training ViT-B/32 on \dsname\ achieves 1.3\%-6.9\% average performance improvement.}
    \vspace{-2mm}
    \label{table:linear_probe}
    \resizebox{\linewidth}{!}{
    \begin{tabular}{cccccccccccccccccccccccc}
        \toprule
        \belowrulesepcolor{lightgray}
         \rowcolor{lightgray} Data Scale & Dataset & \rotatebox[origin=lb]{90}{\smash{Food101}} & \rotatebox[origin=lb]{90}{\smash{CIFAR10}} & \rotatebox[origin=lb]{90}{\smash{CIFAR100}} & \rotatebox[origin=lb]{90}{\smash{Birdsnap}} & \rotatebox[origin=lb]{90}{\smash{SUN397}} & \rotatebox[origin=lb]{90}{\smash{Cars}} & \rotatebox[origin=lb]{90}{\smash{Aircraft}} & \rotatebox[origin=lb]{90}{\smash{DTD}} & \rotatebox[origin=lb]{90}{\smash{Pets}} & \rotatebox[origin=lb]{90}{\smash{Caltech}} & \rotatebox[origin=lb]{90}{\smash{Flowers}} & \rotatebox[origin=lb]{90}{\smash{STL10}} & \rotatebox[origin=lb]{90}{\smash{EuroSAT}} & \rotatebox[origin=lb]{90}{\smash{RESISC45}} & \rotatebox[origin=lb]{90}{\smash{KITTI}} & \rotatebox[origin=lb]{90}{\smash{Country}} & \rotatebox[origin=lb]{90}{\smash{UCF101}} & \rotatebox[origin=lb]{90}{\smash{Memes}} & \rotatebox[origin=lb]{90}{\smash{SST2}} & \rotatebox[origin=lb]{90}{\smash{ImageNet}} & \rotatebox[origin=lb]{90}{\smash{\textbf{Average}}} \\
        \aboverulesepcolor{lightgray}
        \midrule
        \multirow{3}{*}{15M}
         & YFCC & 67.2 & 90.4 & 70.8 & \colorbox{blue_ours}{47.7} & 66.7& 23.8 & 29.7 & 62.4 & 65.7 & 80.1 & \colorbox{blue_ours}{90.0} & 94.7 & 94.9 & 79.4 & \colorbox{blue_ours}{75.4} & \colorbox{blue_ours}{18.4} & 70.8 & 48.6 & 56.2 & 56.7 & 64.5 \\
         & LAION &  71.0 & 93.3 & 78.1 & 41.0 & 66.3 & \colorbox{blue_ours}{76.9} & \colorbox{blue_ours}{43.0} & 71.2 & 74.5 & 87.6 & 88.2 & 93.6 & 95.3 & 82.9 & 72.2 & 13.5 & 75.4 & 55.7 & 57.3 & 59.3 & 69.8 \\
         & \dsname &  \colorbox{blue_ours}{77.1} & \colorbox{blue_ours}{94.5} & \colorbox{blue_ours}{78.7} & 43.4 & \colorbox{blue_ours}{71.4} & 64.7 & 42.7 & \colorbox{blue_ours}{71.3} & \colorbox{blue_ours}{79.9} & \colorbox{blue_ours}{90.0} & 88.2 & \colorbox{blue_ours}{96.4} & \colorbox{blue_ours}{96.2} & \colorbox{blue_ours}{87.2} & 72.4 & 16.7 & \colorbox{blue_ours}{79.9} & \colorbox{blue_ours}{55.7} & \colorbox{blue_ours}{57.7} & \colorbox{blue_ours}{64.0} & \colorbox{blue_ours}{71.4} \\
         \midrule
         \multirow{2}{*}{30M}
         & LAION& 76.1 & 94.5 & 80.0 & 47.4 & 70.3 & \colorbox{blue_ours}{82.3} & \colorbox{blue_ours}{45.9} & \colorbox{blue_ours}{74.7} & 80.3 & 89.8 & 89.5 & 95.6 & 95.5 & 84.5 & 72.6 & 15.2 & 76.6 & \colorbox{blue_ours}{56.2}& \colorbox{blue_ours}{60.0} & 64.3 & 72.6 \\
         & \dsname & \colorbox{blue_ours}{81.2} & \colorbox{blue_ours}{95.4} & \colorbox{blue_ours}{81.8} & \colorbox{blue_ours}{48.4} & \colorbox{blue_ours}{74.5} & 73.4 & 45.2 & 74.2& \colorbox{blue_ours}{84.1} & \colorbox{blue_ours}{91.3} & \colorbox{blue_ours}{90.6} & \colorbox{blue_ours}{97.2} & \colorbox{blue_ours}{96.5} & \colorbox{blue_ours}{89.2} & \colorbox{blue_ours}{74.5} & \colorbox{blue_ours}{19.0} & \colorbox{blue_ours}{82.6} & 55.0 & 56.2 & \colorbox{blue_ours}{68.5} & \colorbox{blue_ours}{73.9} \\
         \midrule
         \multirow{2}{*}{100M}
         & LAION& 80.2 & 95.7 & 82.5 & 51.3 & 73.4 & \colorbox{blue_ours}{85.3} & 46.1 & 75.6 & 83.2 & 91.1 & \colorbox{blue_ours}{92.0} & 96.9 & 95.2 & 85.9 & 68.4 & 17.4 & 80.0 & 57.3& \colorbox{blue_ours}{61.4} & 68.3 & 74.4 \\
         & \dsname & \colorbox{blue_ours}{84.2} & \colorbox{blue_ours}{96.3} & \colorbox{blue_ours}{83.5} & \colorbox{blue_ours}{54.0} & \colorbox{blue_ours}{76.2} & 77.4 & \colorbox{blue_ours}{47.6} & \colorbox{blue_ours}{75.6} & \colorbox{blue_ours}{86.3} & \colorbox{blue_ours}{92.1} & 91.7 & \colorbox{blue_ours}{97.7} & \colorbox{blue_ours}{96.8} & \colorbox{blue_ours}{90.6} & \colorbox{blue_ours}{73.1} & \colorbox{blue_ours}{21.1} & \colorbox{blue_ours}{83.7} & \colorbox{blue_ours}{57.3} & 58.9 & \colorbox{blue_ours}{71.6} & \colorbox{blue_ours}{75.8} \\
        \bottomrule
    \end{tabular}
    }    
\vspace{-2mm}
\end{table*}

%% file: Tables/Zeroshot_classification.tex
\begin{table*}[t!]
    \centering
    \caption{Zero-shot transfer on 20 downstream datasets. Pre-training ViT-B/32 on \dsname\ achieves 2.3\%-14.3\% average performance improvement.}
    \vspace{-2mm}
    \label{table:zeroshot_classfication}
    \resizebox{\linewidth}{!}{
    \begin{tabular}{cccccccccccccccccccccccc}
        \toprule
        \belowrulesepcolor{lightgray}
        \rowcolor{lightgray} Data Scale &  Dataset & \rotatebox[origin=lb]{90}{\smash{Food101}} & \rotatebox[origin=lb]{90}{\smash{CIFAR10}} & \rotatebox[origin=lb]{90}{\smash{CIFAR100}} & \rotatebox[origin=lb]{90}{\smash{Birdsnap}} & \rotatebox[origin=lb]{90}{\smash{SUN397}} & \rotatebox[origin=lb]{90}{\smash{Cars}} & \rotatebox[origin=lb]{90}{\smash{Aircraft}} & \rotatebox[origin=lb]{90}{\smash{DTD}} & \rotatebox[origin=lb]{90}{\smash{Pets}} & \rotatebox[origin=lb]{90}{\smash{Caltech}} & \rotatebox[origin=lb]{90}{\smash{Flowers}} & \rotatebox[origin=lb]{90}{\smash{STL10}} & \rotatebox[origin=lb]{90}{\smash{EuroSAT}} & \rotatebox[origin=lb]{90}{\smash{RESISC45}} & \rotatebox[origin=lb]{90}{\smash{KITTI}} & \rotatebox[origin=lb]{90}{\smash{Country}} & \rotatebox[origin=lb]{90}{\smash{UCF101}} & \rotatebox[origin=lb]{90}{\smash{Memes}} & \rotatebox[origin=lb]{90}{\smash{SST2}} & \rotatebox[origin=lb]{90}{\smash{ImageNet}} & \rotatebox[origin=lb]{90}{\smash{\textbf{Average}}} \\
        \aboverulesepcolor{lightgray}
        \midrule
        \multirow{3}{*}{15M} 
        & YFCC & 36.3 & 74.0 & 40.3 & \colorbox{blue_ours}{19.4} & 41.8 & 2.1 & 2.3 & 12.0 & 19.8 & 59.8 & \colorbox{blue_ours}{48.9} & 87.7 & 21.2 & 20.3 & 23.8 & 5.1 & 27.8 & 47.4 & 50.1 & 32.3 & 33.6 \\
         & LAION & 49.1 & 85.7 & 56.9 & 11.5 & 45.1 & \colorbox{blue_ours}{49.9} & 3.8 & 25.7 & 54.6 & 78.1 & 30.5 & 89.5 & 36.7 & 36.1 & 21.7 & 5.6 & 38.2 & 48.8 & 49.9 & 37.1 & 42.7 \\
         & \dsname & \colorbox{blue_ours}{60.0} & \colorbox{blue_ours}{85.7} & \colorbox{blue_ours}{58.3} & 10.5 & \colorbox{blue_ours}{56.4} & 27.6 & \colorbox{blue_ours}{5.5} & \colorbox{blue_ours}{33.2} & \colorbox{blue_ours}{61.7} & \colorbox{blue_ours}{80.2} & 31.2 & \colorbox{blue_ours}{92.4} & \colorbox{blue_ours}{56.5} & \colorbox{blue_ours}{56.2} & \colorbox{blue_ours}{34.0} & \colorbox{blue_ours}{8.9} & \colorbox{blue_ours}{52.6} & \colorbox{blue_ours}{53.3} & \colorbox{blue_ours}{51.3} & \colorbox{blue_ours}{43.3} & \colorbox{blue_ours}{47.9} \\
         \midrule
         \multirow{2}{*}{30M} 
         & LAION & 58.9 & 85.9 & 63.1 & \colorbox{blue_ours}{17.4} & 54.8 & \colorbox{blue_ours}{61.0} & 4.3 & 36.4 & 65.5& 82.0 & 41.3 & 91.3 & 40.3 & 43.7 & 24.3 & 7.2 & 47.4 & 51.5 & 50.1 & 44.9 & 48.6 \\
         & \dsname & \colorbox{blue_ours}{67.5} & \colorbox{blue_ours}{89.0} & \colorbox{blue_ours}{65.2} & 15.0 & \colorbox{blue_ours}{60.6} & 39.2 & \colorbox{blue_ours}{7.9} & \colorbox{blue_ours}{37.8} & \colorbox{blue_ours}{70.5} & \colorbox{blue_ours}{84.0} & \colorbox{blue_ours}{42.2} & \colorbox{blue_ours}{93.8} & \colorbox{blue_ours}{59.9} & \colorbox{blue_ours}{61.9} & \colorbox{blue_ours}{27.7} & \colorbox{blue_ours}{10.6} & \colorbox{blue_ours}{56.7} & \colorbox{blue_ours}{52.5} & \colorbox{blue_ours}{50.1} & \colorbox{blue_ours}{50.9} & \colorbox{blue_ours}{52.1} \\
         \midrule
         \multirow{2}{*}{100M} 
         & LAION & 68.9 & \colorbox{blue_ours}{90.5} & 68.6 & \colorbox{blue_ours}{23.6} & 60.6 & \colorbox{blue_ours}{68.3} & 7.8 & 41.2 & 74.7 & 87.1 & 47.7 & 94.4 & 45.6 & 53.4 & 23.6 & 10.4 & 54.5 & 51.9 & 53.3 & 52.8 & 53.9 \\
         & \dsname & \colorbox{blue_ours}{73.5} & 89.5 & \colorbox{blue_ours}{68.8} & 20.1 & \colorbox{blue_ours}{65.0} & 48.5 & \colorbox{blue_ours}{10.2} & \colorbox{blue_ours}{46.1} & \colorbox{blue_ours}{76.7} & \colorbox{blue_ours}{87.6} & \colorbox{blue_ours}{48.8} & \colorbox{blue_ours}{94.4} & \colorbox{blue_ours}{69.0} & \colorbox{blue_ours}{65.5} & \colorbox{blue_ours}{24.6} & \colorbox{blue_ours}{12.1} & \colorbox{blue_ours}{60.5} & \colorbox{blue_ours}{52.4} & \colorbox{blue_ours}{54.1} & \colorbox{blue_ours}{56.2} & \colorbox{blue_ours}{56.2} \\
        \bottomrule
    \end{tabular}
    }
\vspace{-2mm}
\end{table*}

%% file: Tables/Zeroshot_retrieval.tex
\begin{table*}[t!]
\centering
\caption{Zero-shot image-text retrieval performance on Flickr30k and MSCOCO. Pre-training CLIP-B/32 on \dsname\ dataset achieves a significant improvement on all metrics.}
\vspace{-2mm}
\label{tab:retrieval}
\resizebox{0.85\linewidth}{!}{
    \centering
    \begin{tabular}{cccccccccccccc}
    \toprule
    \belowrulesepcolor{lightgray}
     \rowcolor{lightgray}  & &\multicolumn{6}{c}{Text Retrieval} & \multicolumn{6}{c}{Image Retrieval} \\
     \rowcolor{lightgray}  & &\multicolumn{3}{c}{Flickr30k} & \multicolumn{3}{c}{MSCOCO} & \multicolumn{3}{c}{Flickr30k} & \multicolumn{3}{c}{MSCOCO} \\
     \rowcolor{lightgray} Data Scale & Dataset& R@1 & R@5 & R@10 & R@1 & R@5 & R@10 & R@1 & R@5 & R@10 & R@1 & R@5 & R@10 \\
    \aboverulesepcolor{lightgray} %
    \midrule \multirow{3}{*}{15M} 
    & YFCC & 37.1 & 64.8 & 75.9 & 21.3 & 45.1 & 57.0 & 23.5 & 47.3 & 58.3 & 13.2 & 32.0 & 43.1 \\
    & LAION & 49.1 & 76.8 & 84.5 & 28.4 & 53.0 & 64.9 & 33.3 & 60.5 & 70.9 & 17.4 & 38.3 & 49.7 \\
    & \dsname & \colorbox{blue_ours}{72.9} & \colorbox{blue_ours}{91.1} & \colorbox{blue_ours}{95.1} & \colorbox{blue_ours}{43.8} & \colorbox{blue_ours}{69.5} & \colorbox{blue_ours}{79.6} & \colorbox{blue_ours}{49.5} & \colorbox{blue_ours}{76.3} & \colorbox{blue_ours}{84.6} & \colorbox{blue_ours}{25.8} & \colorbox{blue_ours}{50.6} & \colorbox{blue_ours}{62.5} \\
    \midrule \multirow{2}{*}{30M} 
    & LAION & 59.6 & 83.5 & 89.8 & 35.9 & 62.4 & 73.2 & 42.4 & 70.1 & 79.4 & 22.1 & 45.5 & 57.6 \\
    & \dsname & \colorbox{blue_ours}{76.0} & \colorbox{blue_ours}{93.3} & \colorbox{blue_ours}{96.9} & \colorbox{blue_ours}{48.2} & \colorbox{blue_ours}{74.6} & \colorbox{blue_ours}{83.0} & \colorbox{blue_ours}{54.0} & \colorbox{blue_ours}{80.0} & \colorbox{blue_ours}{87.6} & \colorbox{blue_ours}{29.5} & \colorbox{blue_ours}{55.2} & \colorbox{blue_ours}{66.9} \\
    \midrule \multirow{2}{*}{100M} 
    & LAION & 67.5 & 87.9 & 93.0 & 43.3 & 68.0 & 78.1 & 50.4 & 77.2 & 85.5 & 27.1 & 52.1 & 63.8\\
    & \dsname & \colorbox{blue_ours}{81.6} & \colorbox{blue_ours}{96.1} & \colorbox{blue_ours}{97.3} & \colorbox{blue_ours}{52.3} & \colorbox{blue_ours}{76.7} & \colorbox{blue_ours}{85.0} & \colorbox{blue_ours}{58.8} & \colorbox{blue_ours}{84.1} & \colorbox{blue_ours}{90.5} & \colorbox{blue_ours}{32.5} & \colorbox{blue_ours}{58.9} & \colorbox{blue_ours}{70.2} \\
    \bottomrule
    \end{tabular}
    }
\vspace{-1mm}
\end{table*}

%% file: Tables/Zeroshot_robustness.tex
\begin{table}[t!]
\centering
\caption{Zero-shot robustness comparison on different IN-1K val set variants. Pre-training CLIP-B/32 on \dsname\ demonstrates superior robustness across all datasets.}
\vspace{-2mm}
\label{tab:robustness}
\resizebox{\linewidth}{!}{
    \begin{tabular}{cccccccc}
        \toprule
        \belowrulesepcolor{lightgray}
        \rowcolor{lightgray} Data Scale  & Dataset& IN-V2 & IN-A & IN-R  & ObjectNet & IN-Sketch & \textbf{Average}\\
        \aboverulesepcolor{lightgray} %
        \midrule \multirow{3}{*}{15M} 
        & YFCC & 27.3 & 12.3 & 20.8 & 25.3 & 6.3 & 18.4  \\
        & LAION & 30.7 & 6.0 & 46.5 & 28.7 & 24.3 & 27.2  \\
        & \dsname & \colorbox{blue_ours}{37.1} & \colorbox{blue_ours}{12.5} & \colorbox{blue_ours}{47.7} & \colorbox{blue_ours}{35.0} & \colorbox{blue_ours}{25.4} & \colorbox{blue_ours}{31.5}  \\
        \midrule \multirow{2}{*}{30M} 
        & LAION & 37.5 & 8.9 & 54.4 & 35.5 & 31.8 & 33.6  \\
        & \dsname & \colorbox{blue_ours}{42.9} & \colorbox{blue_ours}{16.1} & \colorbox{blue_ours}{56.6} & \colorbox{blue_ours}{41.5} & \colorbox{blue_ours}{31.9} & \colorbox{blue_ours}{37.8}  \\
        \midrule \multirow{2}{*}{100M} 
        & LAION & 44.6 & 12.2 & 62.5 & 42.2 & 37.9 & 39.9  \\
        & \dsname & \colorbox{blue_ours}{47.6} & \colorbox{blue_ours}{19.7} & \colorbox{blue_ours}{62.5} & \colorbox{blue_ours}{45.8} & \colorbox{blue_ours}{37.9} & \colorbox{blue_ours}{42.7}  \\
	\bottomrule
    \end{tabular}
    }
\vspace{-3mm}
\end{table}

%% file: sections/4Experiments.tex
\section{Experiments and Results}

\subsection{Implementation Details}
We initially collect 118M interleaved image-text documents from the OBELICS~\cite{Obelics} as our primary data source. We use OFA$_{base}$~\cite{OFA} and RAM++$_{large}$~\cite{RAM_plus_plus} to generate brief captions and semantic tags. To validate the dataset performance, we pre-train standard CLIP supervised by the text randomly selected from the three retrieved realistic texts and one synthetic text, inspired by the LaCLIP~\cite{laclip}. During pre-training, we adopt AdamW~\cite{AdamW} as the optimizer, with a learning rate of 1e-3 and a weight decay of 0.2. The parameters $\beta1$ and $\beta2$ are set to 0.9 and 0.98, respectively. The input image size is 224×224, and the input text sequence length is 77. The temperature parameter $\tau$ is initialized to 0.07. We train 32 epochs with 4096 batch sizes on 8 $\times$ A100 (80G) GPUs. Please refer to the supplementary material for more details.

To validate the effectiveness of the \dsname\ dataset, we compare \dsname\ with the previous datasets across various models and data scales. We compare \dsname15M with the YFCC15M filtered by DeCLIP~\cite{DECLIP}. Following ALIP~\cite{ALIP}, we also compare with LAION15M, LAION30M, and LAION100M~(subset randomly selected from LAION400M).

\begin{figure*}[t!]
    \centering
    \includegraphics[width=\linewidth]{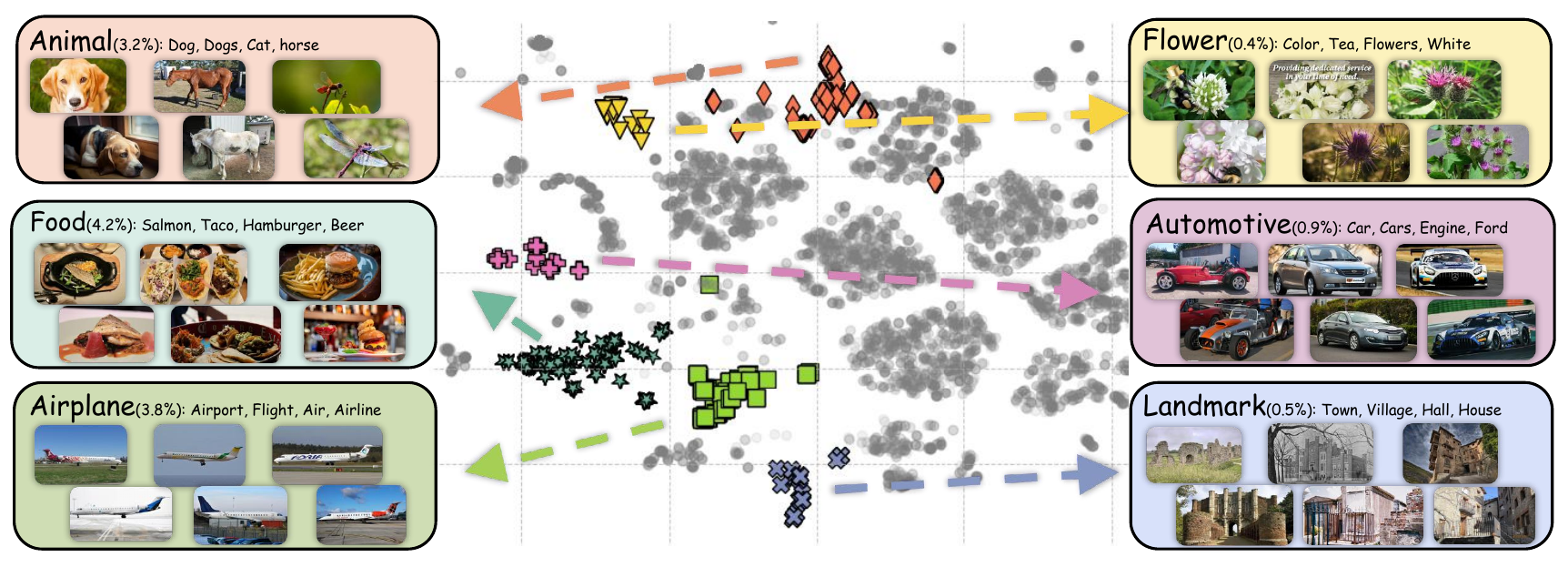}
    \vspace{-6mm}
    \caption{A T-SNE~\cite{tsne} projection of LDA~\cite{LDA} topic cluster from a randomly selected 1M samples from \dsname. \dsname\ encompasses a broad range of everyday topics, e.g., animal, food, airplane, etc. We also display representative central images for each identified topic.}
    \vspace{-2mm}
    \label{fig: dataset_topic_distribution}
\end{figure*}

\subsection{Main Results}
\label{main_result}
\noindent{\bf Linear Probe.} In Table~\ref{table:linear_probe}, we report the linear probe performance of the ViT-B/32 model across 20 downstream datasets. When pretrained at the 15M scale, \dsname15M exceeds YFCC15M on 16 of 20 datasets, achieving an average performance increase of 6.9\%. Additionally, \dsname15M outperforms LAION15M on 18 of 20 datasets, with an average improvement of 1.6\%. With dataset scaling to 30M and 100M, \dsname\ achieves average performance improvements of 1.3\% and 1.4\% over LAION, respectively. These findings underscore the superior CLIP training efficiency of the \dsname\ dataset compared to the widely-used YFCC and LAION datasets across multiple scales.

\noindent{\bf Zero-shot Transfer.} We evaluate the zero-shot transfer performance of the ViT-B/32 model across 20 classification benchmarks using the same prompt templates as SLIP~\cite{mu2022slip}. As indicated in Table~\ref{table:zeroshot_classfication}, \dsname15M surpasses YFCC15M on 18 of 20 datasets, achieving an average performance improvement of 14.3\%. In comparison to LAION15M, \dsname15M excels on 18 of 20 datasets with an average improvement of 5.2\%. Upon expanding the dataset sizes to 30M and 100M, \dsname\ achieves average performance improvements of 3.5\% and 2.3\% compared to LAION, highlighting its efficiency and scalability.

It is important to note that \dsname\ demonstrates a significant decrease in performance on certain datasets, such as Cars and Flowers. This reduction is primarily attributed to the unique data distribution of \dsname, characterized by a scarcity of data for specific concepts, which hampers the model’s ability to effectively learn these concepts. For example, as shown in Figure~\ref{fig: dataset_topic_distribution}, samples related to cars and flowers represent only 0.9\% and 0.4\% of the dataset, respectively.

\noindent{\bf Zero-shot Image-Text Retrieval.} In Table~\ref{tab:retrieval}, we present the zero-shot image-text retrieval performance of the ViT-B/32 model pre-trained on different scales of datasets. \dsname\ achieves superior results across all evaluation metrics. Specifically, \dsname15M improves Recall@1 by 35.8\%\&26\% on Flickr30K~\cite{flickr30k} and by 22.5\%\&12.6\% on MSCOCO~\cite{mscoco_retrieval}. \dsname30M improves Recall@1 by 16.4\%\&11.6\% on Flickr30K~\cite{flickr30k} and by 12.3\%\&7.4\% on MSCOCO~\cite{mscoco_retrieval}. \dsname100M
improves Recall@1 by on 14.6\%\&8.4\% Flickr30K~\cite{flickr30k} and by 9\%\&5.4\% on MSCOCO~\cite{mscoco_retrieval}. This significant enhancement in cross-modal retrieval performance indicates that the \dsname\ dataset effectively improves contrastive vision-language representation learning by incorporating both realistic and synthetic texts, resulting in robust representations and enhanced cross-modal alignment.

\noindent{\bf Zero-shot Robustness.} To further validate the robustness of the \dsname\ dataset, we present the zero-shot robustness~\cite{CLIP} performance across various IN-1K~\cite{ImageNet} val set variants in Table~\ref{tab:robustness}. The results indicate that \dsname\ significantly enhances the robustness of vision-language pre-training models. Specifically, \dsname15M shows a performance increase of 13.1\% and 4.3\% compared to YFCC15M and LAION15M, respectively. When scaled to 30M and 100M, \dsname\ achieves additional improvements of 4.2\% and 2.8\% over LAION. This substantial enhancement in performance is primarily attributable to the utilization of retrieved realistic texts, which surpass the constraints of generative models, and the superior conceptual diversity relative to YFCC and LAION, thus significantly boosting model robustness.

\begin{figure}[!t]
\centering
    \begin{subfigure}{0.22\textwidth}
    \includegraphics[width=1.0\textwidth]{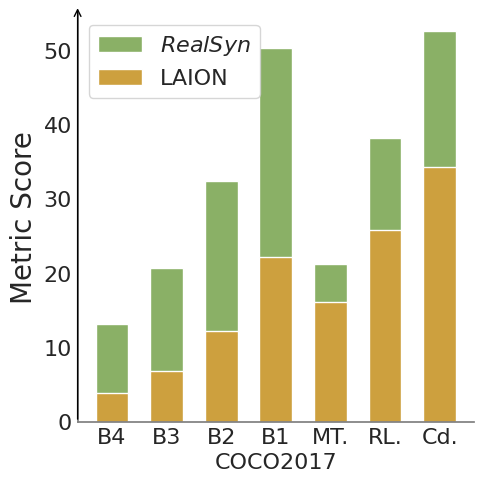}
    \end{subfigure}
    \begin{subfigure}{0.22\textwidth}
    \includegraphics[width=1.0\textwidth]{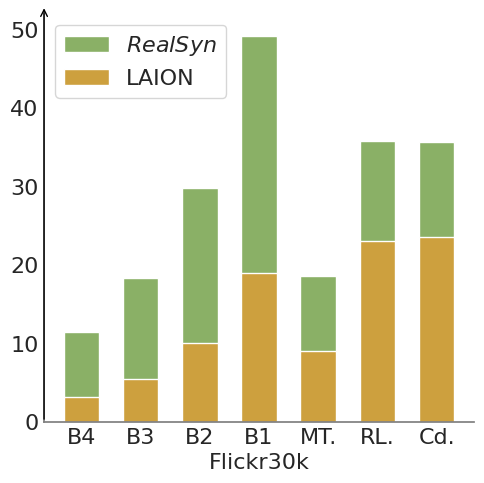}
    \end{subfigure}
\vspace{-3mm}
\caption{Image captioning comparisons on COCO2017 and Flickr30k. B4, MT., RL. and Cd. represent the metric of BLEU~\cite{BLEU}, METEOR~\cite{METEOR}, ROUGE-L~\cite{METEOR}, and Cider~\cite{cider}.}
\vspace{-5mm}
\label{image_captioning}
\end{figure}

\noindent{\bf Image Captioning via MLLM.} In Figure~\ref{image_captioning}, we present the image captioning performance of the LLaVA-1.5~\cite{llava1.5} trained using different datasets (LAION v.s. \dsname). Initially, we first map visual features into the textual domain using the initial 558k dataset from LLaVA-1.5. We then develop an image captioning dataset from both LAION and \dsname\ for instruction tuning. Specifically, we select 1M samples randomly from each dataset and train over two epochs. As depicted in Figure~\ref{image_captioning}, \dsname\ significantly outperforms LAION in all evaluation metrics on both the COCO2017~\cite{mscoco_retrieval} and Flickr30k~\cite{flickr30k} benchmarks. This notable performance enhancement confirms the higher quality and better image-text alignment of the \dsname\ dataset.

\begin{figure}[t!]
\centering
\begin{subfigure}{\linewidth}
{
\includegraphics[height=0.48\textwidth]{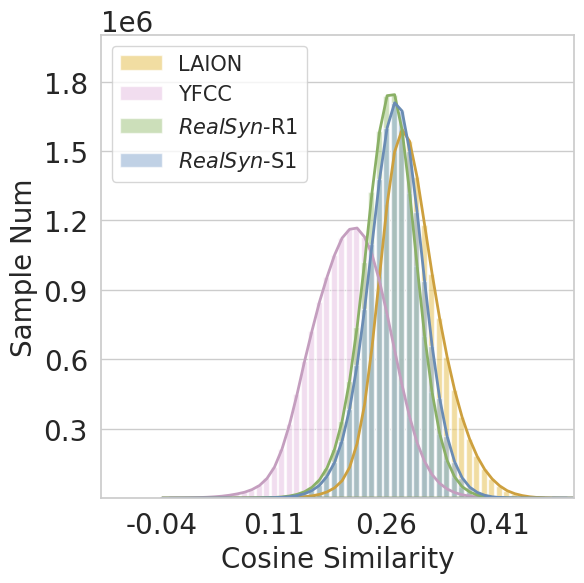}}
{
\includegraphics[height=0.48\textwidth]{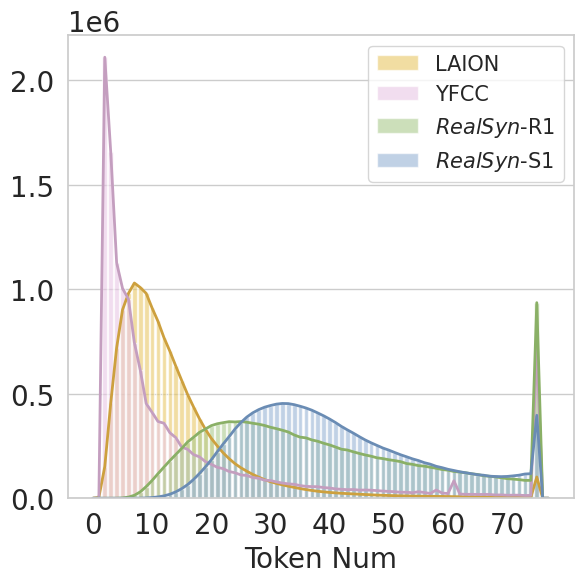}}
\caption{Richness assessment comparison}
\label{similarity_token}
\end{subfigure}

\begin{subfigure}{\linewidth}
\centering
\includegraphics[width=\linewidth]{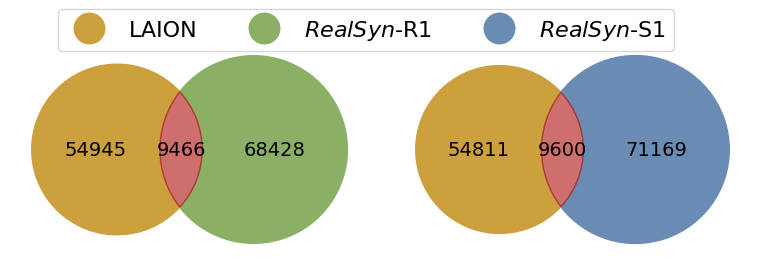}
\vspace{-7mm}
\caption{Diversity assessment comparison}
\label{fig:diversity}
\end{subfigure}
\vspace{-5mm}
\caption{The richness assessment and diversity assessment on different datasets. \dsname-R1: the most relevant retrieved realistic text. \dsname-S1: the semantic augmented synthetic text based on \dsname-R1.}
\vspace{-3mm}
\end{figure}

%% file: sections/5Ablation_study.tex
\section{Analysis}

\begin{figure}[t!]
\centering
\includegraphics[width=\linewidth]{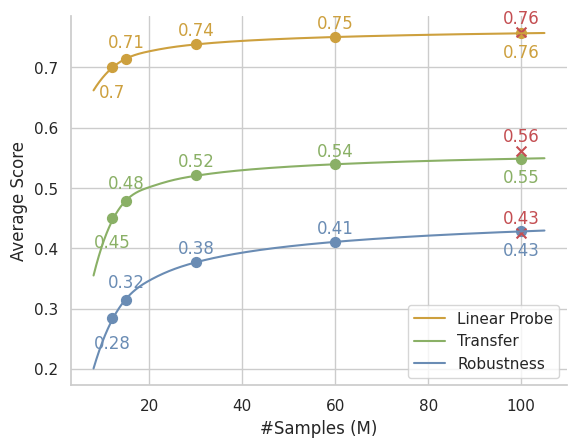}
\vspace{-7mm}
\caption{Data Scaling Analysis. We pretrain ViT-B/32 on \dsname\ in different data scales. \textcolor{red}{$\times$} represents the results predicted using our data scaling law.}
\label{fig:data_scaling_analysis}
\vspace{-3mm}
\end{figure}

\subsection{Statistics Analysis} 
\noindent{\bf Topic-based Assessment.} Following MMC4~\cite{MMC4}, we ran LDA~\cite{LDA} on random sampling 1M image-realistic text pairs with 30 topics. Figure~\ref{fig: dataset_topic_distribution} presents the proportions and examples for six topics: animal, food, airplane, flower, automotive, and landmark. Notably, the dataset contains minimal samples related to ``flower'' and ``automotive'' topics, representing merely 0.4\% and 0.9\% of the total, respectively. This paucity of examples hinders the model's ability to sufficiently learn these concepts, thereby compromising its performance in the linear probe and zero-shot transfer evaluations on the Flowers and Cars datasets.

\noindent{\bf Richness Assessment.} Figure~\ref{similarity_token} presents image-text similarity and text token distribution of 15M samples from YFCC15, LAION, \dsname-R1 (the most relevant retrieved realistic text), and \dsname-S1 (the semantic augmented synthetic text based on \dsname-R1). Compared to datasets sourced from the Internet, \dsname\ demonstrates robust similarity metrics even after the removal of OCR data. This effectively explains the significant performance enhancement observed in the CLIP model trained on \dsname\ for image-text retrieval tasks.  Moreover, both the retrieved realistic texts and synthetic texts contain a larger quantity of words, which can provide a richer textual context that enhances vision-language representation learning.

\noindent{\bf Diversity Assessment.} The \dsname\ is constructed based on real-world interleaved image-text documents, which encompasses a wide array of diverse information. Following previous work~\cite{lai2024revisit}, we randomly select 0.2M samples to calculate the number of unique entities in the caption to assess the data diversity of different datasets. As depicted in Figure~\ref{fig:diversity}, both the retrieved realistic texts and image semantic augmented synthetic texts exhibit a higher number of distinct entities. Such diversity enriches the dataset, facilitating the model's acquisition of comprehensive knowledge and enhancing both performance and robustness.

\noindent{\bf Data Scaling Analysis.} We present the data scaling law~\cite{scaling_law} derived from our \dsname\ dataset, justifying its scalability over samples.
Specifically, we conduct a series of visual-language pre-trainings with proposed datasets ranging from 12M to 60M, and fit each performance metric to the inverse of logarithmic functions with respect to the number of millions of training samples $x$. Based on the fitting results from these preliminary experiments, we extrapolate each performance scaling law to $100M$ samples, and validate their predicted scaling trends with our \dsname$100M$ dataset as shown in Figure~\ref{fig:data_scaling_analysis}. Notably, as indicated by the coefficients shown in Eq.~\ref{eq:data_scaling_law}, these performance laws also likely suggest an upper bound of model capability that a ViT-B/32 could possibly reach through our proposed visual-language pre-training paradigm with multimodal interleaved documents:
\begin{equation}
\begin{aligned}
    \text{Linear Probe:} \ \mathcal{L}(x) &\approx \frac{-0.21}{log(x-4.23)} + 0.80 \\
    \text{Transfer:} \ \mathcal{L}(x) &\approx \frac{-0.30}{log(x-5.68)} + 0.62 \\
    \text{Robustness:} \ \mathcal{L}(x) &\approx \frac{-0.60}{log(x-3.17)} + 0.56
\end{aligned}
\label{eq:data_scaling_law}
\end{equation}

\begin{figure}[t!]
\centering
\includegraphics[width=1\linewidth]{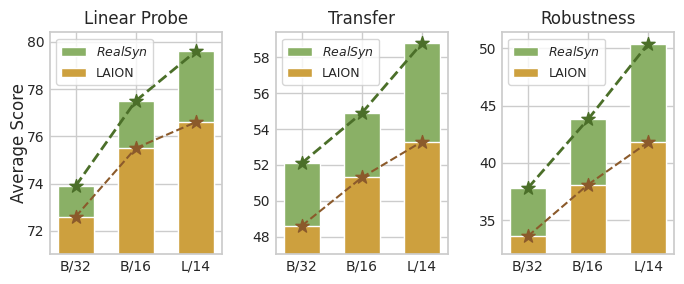}
\vspace{-5mm}
\caption{\textbf{Model scaling capability.} We compare the models pre-trained on LAION30M and \dsname30M.}
\label{model_scale}
\end{figure}

\noindent{\bf Model Scaling Analysis.} In Section~\ref{main_result}, \dsname\ exhibits superior performance across various data scales. To further explore the model scaling capability, we present the downstream task performance of three models in Figure~\ref{model_scale}. Notably, compared to LAION, \dsname\ demonstrates steeper slopes in performance curves across linear probe, zero-shot transfer, and robustness, indicative of its superior model scaling capabilities.

\input{Tables/Concept_balance}

\input{Tables/Ablation_input_texts}

\noindent{\bf Extension to Pure Image.} 
To further extend our transformation paradigm for pure images, we conduct experiments on ImageNet~\cite{ImageNet}. Initially, we retrieve semantically relevant realistic texts for each ImageNet image from our sentence database and generate image semantic augmented synthetic texts. Then, we pre-train ResNet50~\cite{he2016deep} supervised by the text randomly selected from the retrieved realistic texts and synthetic texts. Comparative analysis with SimCLR~\cite{chen2020simple} under identical conditions shows a linear probe average performance enhancement of 2.1\% across 12 datasets using our constructed data. Detailed experimental results are provided in the supplementary material.

\subsection{Ablation Study}

\begin{figure}[t!]
\centering
\includegraphics[width=\linewidth]
{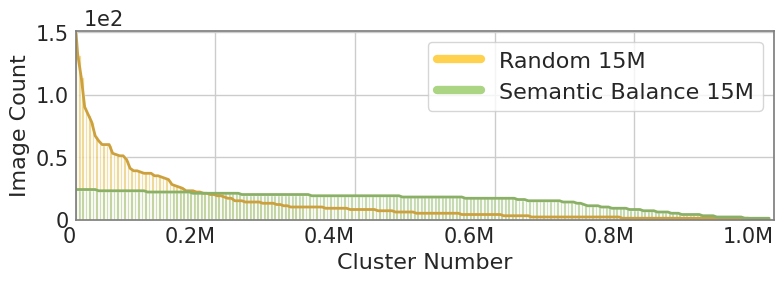}
\vspace{-5mm}
\caption{Clustering distribution of 15M data obtained from random sampling and semantic balance sampling.}
\label{fig:clustering_distribution}
\vspace{-4mm}
\end{figure}

\noindent{\bf Ablation on Semantic Balance Sampling.} To demonstrate the efficacy of our proposed semantic balance sampling method, we contrast it with random sampling. As shown in Table~\ref{tab:concept_balance}, semantic balance sampling achieves performance improvements of 0.7\%, 1.1\%, and 1.0\% in linear probe accuracy, zero-shot transfer, and robustness, respectively. Besides, we visualize the data distribution of 15M samples clustered into 1M centers using different sampling strategies. As shown in Figure~\ref{fig:clustering_distribution} reveals that semantic balance sampling results in a smoother distribution, thereby enhancing the learning of long-tail concepts.

\noindent{\bf Ablation on Realistic Texts and Synthetic Texts.} We conduct ablation studies to assess the impact of varying quantities of realistic and synthetic texts on the performance of the CLIP-B/32 model. As shown in Table~\ref{tab:ablation-text}, incrementally increasing the amount of realistic text from one to three enhances model performance, attributed to improved text augmentation that integrates extensive real-world knowledge. However, further increasing this quantity from three to five slightly diminishes performance due to information saturation and the introduction of noise. Conversely, augmenting the number of synthetic texts from one to five incrementally degrades performance, reflecting increased noise introduction. Notably, training with only realistic texts significantly boosts performance, achieving a 71.2\% accuracy compared to 69.8\% with the LAION15M dataset, underscoring the vital role of real-world knowledge in advancing vision-language representation learning.

\noindent{\bf Ablation on the Combination of the Realistic Texts and Synthetic Texts.} Table~\ref{tab:ablation-text} presents the results of ablation experiments on text augmentation using different text types. Introducing image semantic augmented synthetic text, which supplements fine-grained visual semantic information, leads to performance enhancements of 0.2\%, 1.1\%, and 0.8\% in linear probe accuracy, zero-shot transfer, and zero-shot robustness, respectively, compared to using only retrieved realistic texts.

\definecolor{fig2_green}{HTML}{419A23}
\begin{figure}[!t]
\centering
\includegraphics[width=\linewidth]{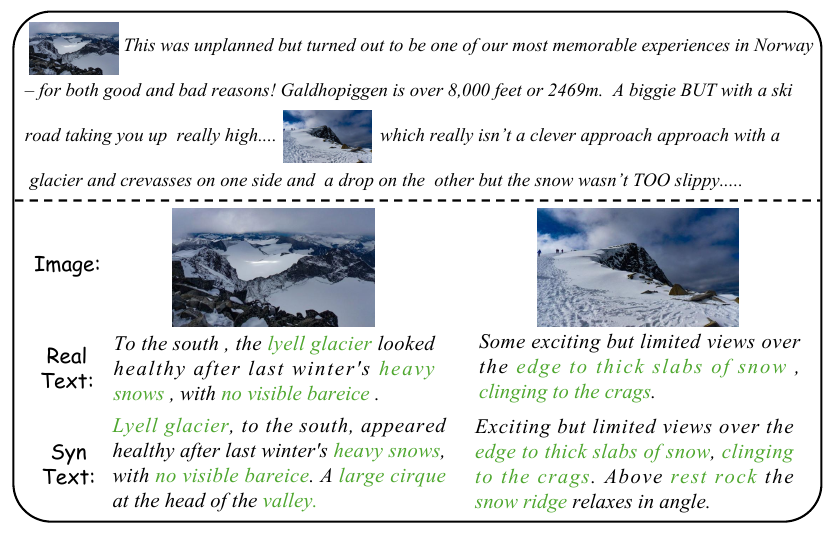}
\vspace{-5mm}
\caption{Visualization of the raw interleaved document, the retrieved realistic text, and synthetic text. Image semantic-related information is highlighted in \textcolor{fig2_green}{green}.}
\label{fig:visualization_example}
\vspace{-4mm}
\end{figure}

\noindent{\bf Case Study.} In Figure~\ref{fig:visualization_example}, we present the visualization of retrieved realistic text and synthetic text obtained from an interleaved image-text document using our proposed transformation paradigm. Both realistic and synthetic texts contain extensive descriptive information consistent with the image semantics, such as ``lyell glacier'', ``crags'', and ``valley''. We provide more visualizations in the supplementary material.

%% file: Tables/Concept_balance.tex
\definecolor{dt}{gray}{0.7}
\begin{table}[t!]
\centering
\caption{Comparison of semantic balance sampling and random sampling on the 15M dataset.}
\vspace{-2mm}
\label{tab:concept_balance}
\resizebox{\linewidth}{!}{
    \begin{tabular}{ccccc}
        \toprule
        \belowrulesepcolor{lightgray}
         \rowcolor{lightgray} Model  & \shortstack{Dataset}  & \shortstack{Linear probe\\ Avg} & \shortstack{Transfer\\ Avg} & \shortstack{Robustness\\ Avg} \\ 
        \aboverulesepcolor{lightgray}
        \midrule
        \multirow{4}{*}{CLIP-B/32} 
        & \color{dt} YFCC  & \color{dt} 64.5 & \color{dt} 33.6 & \color{dt} 18.4  \\
        & \color{dt} LAION  & \color{dt} 69.8 & \color{dt} 42.7 & \color{dt} 27.2  \\
        & \dsname-Random  & 70.7 & 46.8 & 30.5  \\
        & \dsname-Balance  & \colorbox{blue_ours}{71.4} & \colorbox{blue_ours}{47.9} & \colorbox{blue_ours}{31.5} \\
	\bottomrule
    \end{tabular}
    }
\vspace{-3mm}
\end{table}

%% file: Tables/Ablation_input_texts.tex
\begin{table*}[t!]
\centering
\caption{Ablation study examining the quantity and combinations of retrieved realistic texts and corresponding image semantic augmented synthetic texts within a 15M-scale dataset. $T^k_{r}$: the $k$-th retrieved semantic relevant realistic text. $T^k_{s}$: the image semantic augmented synthetic text for $T^k_{r}$.}\label{tab:ablation-text}
\vspace{-2mm}
\begin{minipage}{0.27\textwidth}
\centering
\resizebox{\linewidth}{!}{
    \begin{tabular}{cccccc}
    \toprule
    \belowrulesepcolor{lightgray}
     \rowcolor{lightgray} $T^1_{r}$  & $T^2_{r}$  & $T^3_{r}$  & $T^4_{r}$  & $T^5_{r}$  & \shortstack{Linear probe\\ Avg} \\
    \aboverulesepcolor{lightgray}
    \midrule
    \Checkmark & & & & & 70.3 \\
    \Checkmark &\Checkmark & & &  &  71.0  \\
    \Checkmark &\Checkmark &\Checkmark & &  & \colorbox{blue_ours}{71.2}  \\
    \Checkmark &\Checkmark &\Checkmark &\Checkmark &  & 70.9   \\
    \Checkmark &\Checkmark &\Checkmark &\Checkmark &\Checkmark  & 70.6  \\
    \bottomrule
    \end{tabular}
 } 
\end{minipage}
\begin{minipage}{0.27\textwidth}
\centering
\resizebox{\linewidth}{!}{
    \begin{tabular}{cccccc}
    \toprule
    \belowrulesepcolor{lightgray}
     \rowcolor{lightgray} $T^1_{s}$  & $T^2_{s}$  & $T^3_{s}$  & $T^4_{s}$  & $T^5_{s}$  & \shortstack{Linear probe\\ Avg} \\
    \aboverulesepcolor{lightgray}
    \midrule
    \Checkmark & & & & & \colorbox{blue_ours}{70.2} \\
    \Checkmark &\Checkmark & & &  &  70.0  \\
    \Checkmark &\Checkmark &\Checkmark & &  & 69.9  \\
    \Checkmark &\Checkmark &\Checkmark &\Checkmark &  & 69.4  \\
    \Checkmark &\Checkmark &\Checkmark &\Checkmark &\Checkmark  & 69.1  \\

    \bottomrule
    \end{tabular}
 } 
\end{minipage}
\begin{minipage}{0.45\textwidth}
\centering
\resizebox{\linewidth}{!}{
     \begin{tabular}{ccccccc}
        \toprule
        \belowrulesepcolor{lightgray}
        \rowcolor{lightgray} $T^{1}_{r}$ & $T^{2}_{r}$ & $T^{3}_{r}$ & $T^{1}_{s}$  & \shortstack{Linear probe\\ Avg} & \shortstack{Transfer\\ Avg} & \shortstack{Robustness\\ Avg} \\
        \aboverulesepcolor{lightgray}
        \midrule
         \Checkmark &  & &  & 70.3 & 42.4 & 25.7  \\
         \Checkmark & \Checkmark & \Checkmark &  & 71.2 & 46.8 & 30.7 \\
          &  & & \Checkmark & 70.2 & 39.7 & 24.0 \\
         \Checkmark & \Checkmark & \Checkmark & \Checkmark & \colorbox{blue_ours}{71.4} & \colorbox{blue_ours}{47.9} & \colorbox{blue_ours}{31.5} \\
	\bottomrule
    \end{tabular}
 } 
\end{minipage}
\hfill
\vspace{-2mm}
\end{table*}

%% file: sections/6Conclusion.tex
\section{Conclusion}
This paper explores two open-ended questions: 1) How to utilize multimodal interleaved documents for CLIP training. 2) How to effectively leverage both realistic and synthetic texts to enhance CLIP performance. To this end, we first establish a Real-World Data Extraction pipeline to extract high-quality images and texts. Then we design a hierarchical retrieval method to efficiently associate each image with multiple semantically relevant texts. To further enhance fine-grained visual information, we propose an image semantic augmented generation module for synthetic text production. Furthermore, we employ a semantic balance sampling strategy to improve dataset diversity, enabling better learning of long-tail concepts. Based on these innovations, we present \textit{RealSyn}, a dataset driven by both real and synthetic texts with three sizes: 15M, 30M, and 100M. We compare our dataset with other widely used datasets of equivalent scale for CLIP training. Models pre-trained on RealSyn consistently achieve state-of-the-art performance across various downstream tasks. Furthermore, extensive experiments confirm that \textit{RealSyn} significantly enhances contrastive vision-language representation learning and demonstrates robust scalability. We hope our work provides insights into vision-language representation learning.

%% file: sections/Xappendix.tex
\appendix

This supplementary material introduces the experiment settings and the instruction prompt we used in Section~\ref{sec: detail experiment settings}. Then, we introduce the downstream datasets, detailed model scaling results, comparison with the rewrite datasets, and analyze the combination \dsname\ with LAION and detailed results on pure image in Section~\ref{sec: detail external results}. We further analyze our proposed \dsname\ dataset by comparison with current datasets and visualize examples in Section~\ref{sec: further analysis of data}. Finally, we discuss the limitations of this paper in Section~\ref{sec: limitation}.

\section{Detail Experiment Settings}
\label{sec: detail experiment settings}

\subsection{Experiment Settings}
In Table~\ref{tab:hyperparams}, we present the detailed settings used in training CLIP.
\input{Tables_supp/Experimental_setting}

\subsection{Detail Instruction Prompt}
The prompt we used for ChatGPT to construct the 100K instruction dataset is present in the following:
\\
\\
\emph{"Please merge the information from the given raw text and the synthetic caption with the help of the highly relevant detection tags. The raw caption offers detailed real-world information, yet it suffers from flaws in sentence structure and grammar. The synthetic caption exhibits impeccable sentence structure but often lacks in-depth real-world details and may contain false information. The highly relevant detection tags are provided to enrich the semantic information of the raw caption, while some are redundant and noisy. You are a great information integration and summary expert, you are also good at enriching semantic information. Ensure a well-structured sentence while retaining the detailed real-world information provided in the raw caption. Avoid simply concatenating the sentences and avoid adding external information to describe. Correct and simplify sentences finally. Raw caption:\textless raw caption\textgreater, synthetic caption:\textless synthetic caption\textgreater, and highly relevant detection tags:\textless detection tags\textgreater".}

\section{Detail External Results}
\label{sec: detail external results}

\subsection{Downstream Datasets}

To demonstrate the performance of CLIP trained on \dsname, we compared the linear probe results of CLIP trained on \dsname, YFCC~\cite{YFCC100M}, and LAION~\cite{laion400M} across 20 datasets. These datasets
include Food101~\cite{bossard2014food}, 
CIFAR10~\cite{krizhevsky2009learning}, 
CIFAR100~\cite{krizhevsky2009learning}, 
Birdsnap~\cite{berg2014birdsnap},
SUN397~\cite{xiao2010sun},
Stanford Cars~\cite{KrauseStarkDengFei-Fei_3DRR2013},
FGVC Aircraft~\cite{maji2013fine},
DTD~\cite{cimpoi2014describing},
Pets~\cite{parkhi2012cats}, 
Caltech101~\cite{fei2004learning},
Flowers102~\cite{nilsback2008automated},
SLT10~\cite{coates2011analysis},
EuroSAT~\cite{helber2019eurosat},
RESISC45~\cite{cheng2017remote},
KITTI~\cite{geiger2012we},
Country211~\cite{CLIP},
UCF101~\cite{soomro2012ucf101},
Hateful Memes~\cite{kiela2020hateful},
SST2~\cite{CLIP}, and
ImageNet~\cite{ImageNet}. Details on each dataset and the corresponding evaluation metrics are provided in Table~\ref{linearprobedatasets}.

\input{Tables_supp/linear_probe_datasets}

\subsection{Detailed Model Scaling Results}

\input{Tables_supp/Data_scaling_linear}
\noindent{\bf Linear Probe.} In Table~\ref{table:linear_probe_30M_supp}, we present the detailed linear probe results of different scale CLIP models trained on the 30M dataset. The ViT-L/14 trained on \dsname30M achieves an average performance improvement of 3.0\% across 20 datasets compared to the model trained on the LAION30M.

\input{Tables_supp/Data_scaling_transfer}

\noindent{\bf Zero-shot Transfer.} As shown in Table~\ref{table:zeroshot_classfication_30M_supp}, ViT-L/14 trained on our proposed \dsname30M outperforms LAION30M on 18 of 20 downstream datasets and achieves an average improvement of 5.5\%.

\input{Tables_supp/Data_scaling_robustness}

\noindent{\bf Zero-shot Robustness.} We present the detailed zero-shot robustness performance in Table~\ref{tab:robustness_30M_supp}, compared with LAION30M, the model trained on \dsname30M boosts average robustness performance by 8.6\% on ViT-L/14.

\input{Tables_supp/CMP_DreamLIP}
\subsection{Comparison with the Rewrite Dataset.} 
Due to LaCLIP~\cite{laclip} and Capsfusion~\cite{yu2024capsfusion} do not provide the model weights trained on the same scale dataset as ours, We compared the same scale model ViT-B/16 pre-trained on the \dsname\ with DreamLIP~\cite{DreamLIP}.

As shown in Table~\ref{tab:DreamLIP_cmp}, ViT-B/16 pre-trained on \dsname15M can achieve 1.3\% and 8.3\% average improvement compared to DreamLIP proposed 15M Rewriting datasets based on YFCC15M~\cite{YFCC100M}. Furthermore, compared with DreamLIP's proposed 30M, which combines YFCC15M, CC12M~\cite{CC12M} and CC3M~\cite{CC12M} dataset, ViT-B/16 training on \dsname30M can also outperform it in the linear probe and zero transfer. This significant improvement indicates that realistic knowledge is important for the CLIP model.

\subsection{Combine \dsname\ with LAION}
As shown in Table~\ref{tab: Data_Scaling_Comparison}, integrating \dsname15M with LAION15M yields average improvements of 0.6\% in linear probe, 0.2\% in zero-shot transfer across 20 datasets, and a robustness increase of 0.6\% compared to LAION30M. Comparing \dsname30M with LAION30M, the former shows significant performance gains, with average improvements of 1.3\% in linear probe, 3.5\% in zero-shot transfer, and a robustness enhancement of 4.2\%. The experimental results demonstrate that \dsname\, when combined with existing datasets, achieves significant performance improvements, while also validating the robustness and extensibility of \dsname.

\input{Tables/laion_RS_merge}

\input{Tables_supp/SimCLR_comparison}
\input{Tables_supp/Dataset_comparison}

\subsection{Detailed Results on Pure Image}
To further extend our method for pure images, we conduct experiments on ImageNet~\cite{ImageNet}. For each image, we retrieve three semantically relevant real-world sentences from our pre-constructed sentence database and generate a single semantically augmented synthetic caption based on the top retrieved text. Following SimCLR~\cite{chen2020simple}, we utilize 4096 batch size and pre-train ResNet50~\cite{he2016deep} for 90 epochs supervised by the text randomly selected from the three retrieved realistic texts and one synthetic text.

As shown in Table~\ref{sim_clr_comparison}, compared with SimCLR~\cite{chen2020simple} under the same conditions, the model trained on our constructed image-text pairs shows an average performance improvement of 2.1\% across 12 downstream datasets. The results demonstrate that our method can effectively transform pure images into high-quality image-text pairs through retrieval and generation for vision-language pre-training.

\section{Further Analysis of \dsname}
\label{sec: further analysis of data}

\subsection{Compare with Existing Datasets}
In Table~\ref{public_datasets_comparison}, we compare our proposed \dsname \ dataset with existing widely used large-scale image-text pre-training datasets. Compared to previous datasets, our proposed \dsname\ dataset provides four textual descriptions for each image, with an average token length of 36-40, significantly higher than LAION400M and YFCC15M. Furthermore, unlike previous datasets, the \dsname\ dataset is sourced from real-world interleaved image-text documents and includes both realistic and synthetic texts, thereby expanding the scope for future research exploration.

\subsection{Visualization of Examples}

\definecolor{fig2_green}{HTML}{419A23}
\definecolor{fig2_red}{HTML}{B00003}
\begin{figure*}[t]
    \centering
    \includegraphics[width=0.95\linewidth]{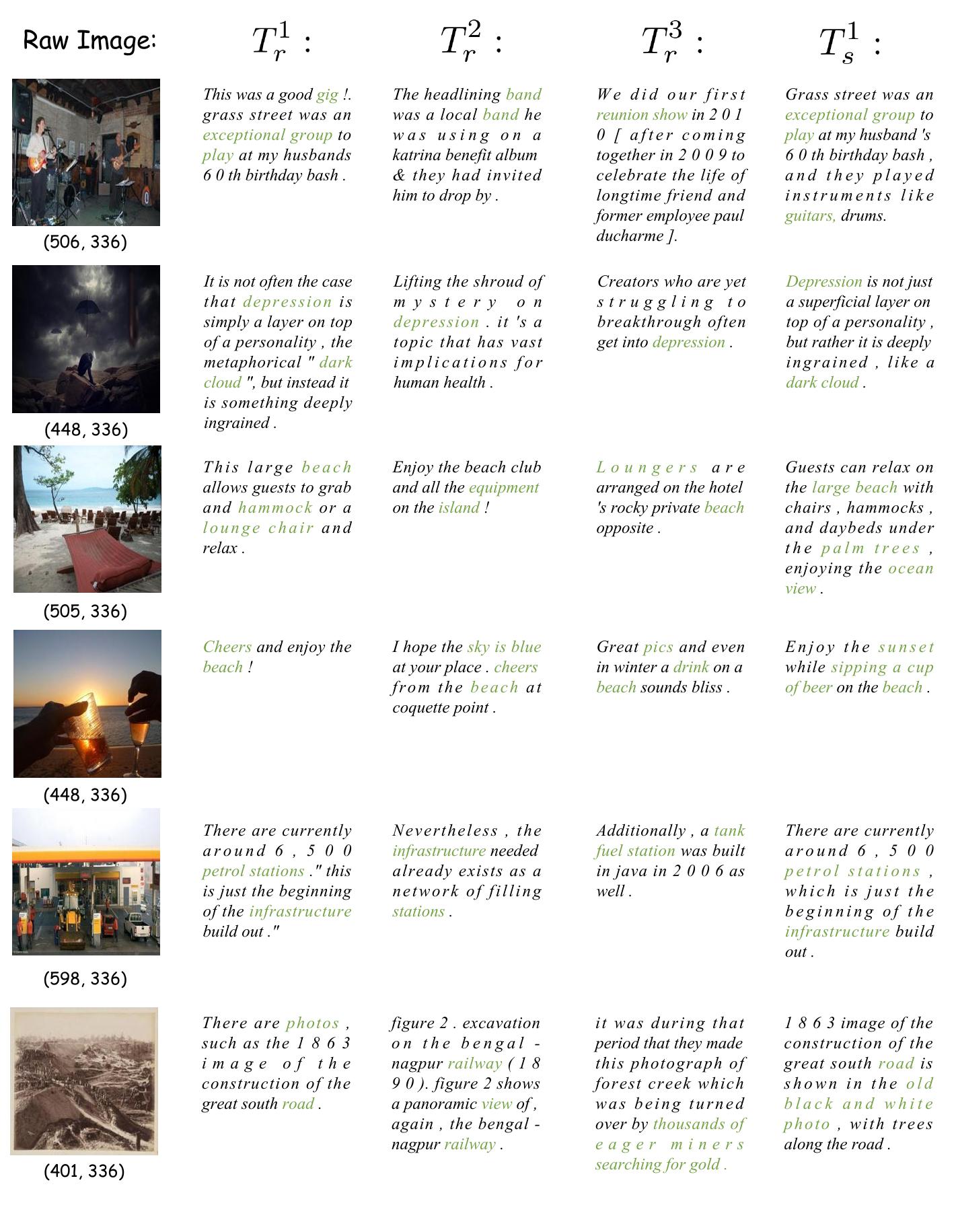}
    \vspace{-8mm}
    \caption{Visualization of image-text pairs in our proposed \dsname\ dataset. $T^{k}_{r}$: the $k$-th retrieved realistic text. $T^{k}_{s}$: the image semantic augmented synthetic text for $T^{k}_{r}$. Image semantic-related information is highlighted in \textcolor{fig2_green}{green}.}
    \label{visualization_example1}
\end{figure*}

\begin{figure*}[t]
    \centering
    \includegraphics[width=0.95\linewidth]{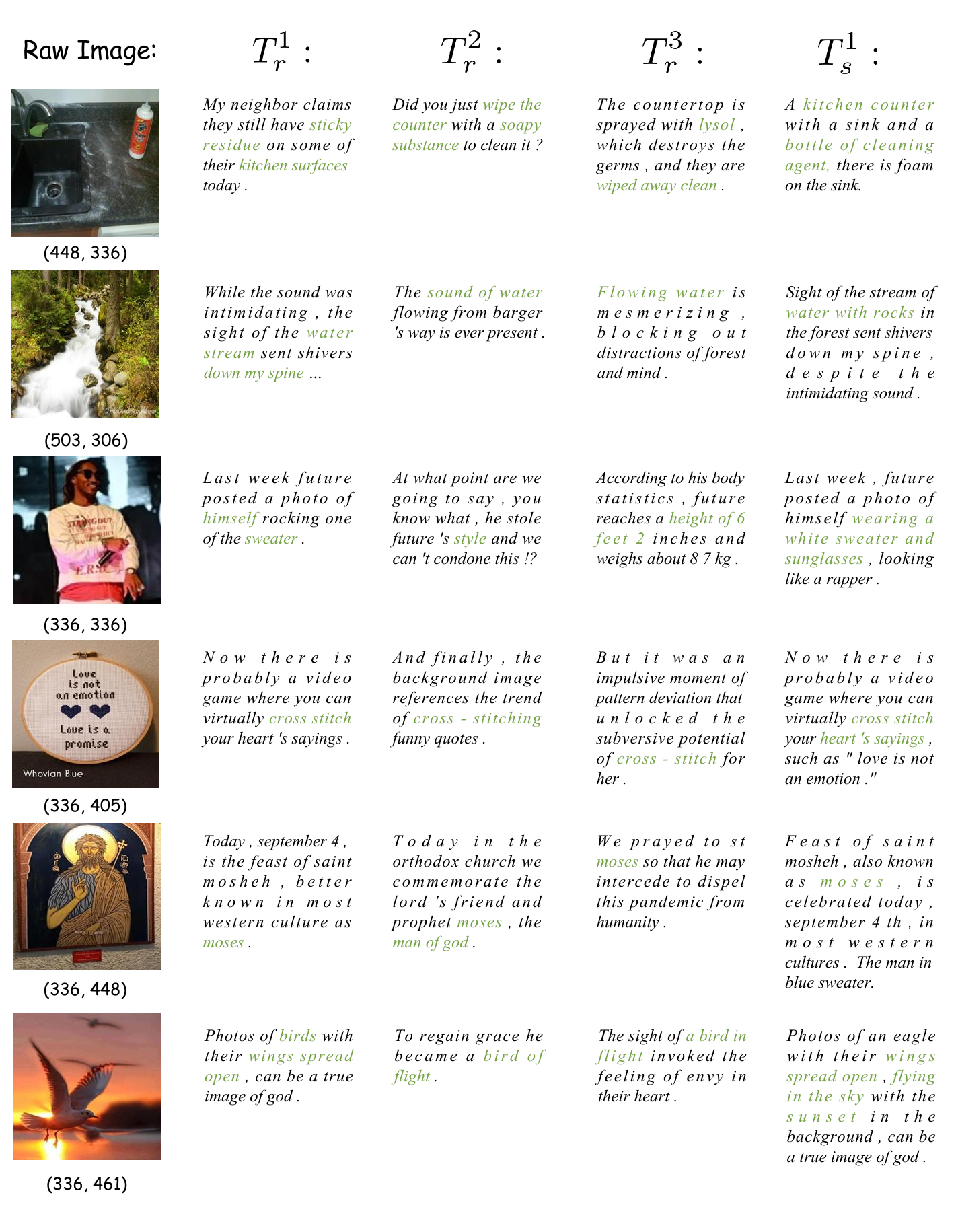}
    \vspace{-8mm}
    \caption{Visualization of image-text pairs in our proposed \dsname\ dataset. $T^{k}_{r}$: the $k$-th retrieved realistic text. $T^{k}_{s}$: the image semantic augmented synthetic text for $T^{k}_{r}$. Image semantic-related information is highlighted in \textcolor{fig2_green}{green}.}
    \label{visualization_example2}
\end{figure*}

In Figure~\ref{visualization_example1} and Figure~\ref{visualization_example2}, we visualize additional image-text pairs randomly selected from our proposed \dsname\ dataset. $T^{k}_{r}$ is the $k$-th retrieved semantically relevant real-world sentence and $T^{k}_{s}$ is the semantic augmentic caption for $T^{k}_{r}$. We also highlighted the image semantic-related information in \textcolor{fig2_green}{green} and marked the image size below the image.

\section{Limitations}
\label{sec: limitation}
To provide more fine-grained visual information, this study employs vision expert models in conjunction with a Large Language Model (LLM) to generate synthetic text. Considering inference costs and efficiency, further exploration of Modified Large Language Models (MLLM) for synthetic text creation is suggested for the community. Additionally, constrained by computational resources, this paper constructs a 100M-scale \dsname\ dataset exclusively using OBELICS~\cite{Obelics}. Notably, the transformation paradigm presented here is directly applicable to other multimodal document datasets, including MMC4~\cite{MMC4} and OmniCorpus~\cite{omnicorpus}.

%% file: Tables_supp/Experimental_setting.tex
\begin{table}[h!]
    \centering
    \caption{Hyperparameters used for CLIP pre-training.}
    \resizebox{0.8\linewidth}{!}{
     \begin{tabular}{l|c}
        \toprule Hyperparameter & Value \\
        \midrule
        Initial temperature & $0.07$ \\
        Weight decay & $0.2$ \\
        Batch size & 4096 \\
        Learning rate & 0.001 \\
        Learning rate scheduler & OneCycleLR \\
        Pct start & 0.1 \\
        Training epochs & 32  \\
        GPU & $8 \times $A100 \\
        Adam $\beta_{1}$ & $0.9$ \\
        Adam $\beta_{2}$ & $0.98$ \\
        Adam $\epsilon$  & $10^{-6}$ \\
        \bottomrule
    \end{tabular}
    }
\centering
\label{tab:hyperparams}
\vspace{-3mm}
\end{table}

%% file: Tables_supp/linear_probe_datasets.tex
\begin{table}[h!]
\centering
\caption{List of linear probe datasets with the data distribution and evaluation metrics.}
\label{linearprobedatasets}
\vspace{-2mm}
\resizebox{0.95\linewidth}{!}{
\begin{tabular}{lcccr}
\toprule
\multicolumn{1}{l}{Dataset} & \multicolumn{1}{c}{Classes} & \multicolumn{1}{c}{Train size} & \multicolumn{1}{c}{Test size} & \multicolumn{1}{c}{Evaluation metric} \\
\midrule
Food101                        & 102                             & 75,750                             & 25,250                            & accuracy                                  \\
CIFAR10                        & 10                              & 50,000                             & 10,000                            & accuracy                                  \\
CIFAR100                       & 100                             & 50,000                             & 10,000                            & accuracy                                  \\
Birdsnap                        & 500                             & 42,138                             & 2,149                             & accuracy                                  \\
SUN397                          & 397                             & 19,850                             & 19,850                            & accuracy                                  \\
Cars                   & 196                             & 8,144                              & 8,041                             & accuracy                                  \\
Aircraft                   & 100                             & 6,667                              & 3,333                             & mean per class                            \\
DTD           & 47                              & 3,760                              & 1,880                             & accuracy                                  \\
Pets                & 37                              & 3,680                              & 3,669                             & mean per class                            \\
Caltech101                     & 101                             & 3,000                              & 5,677                             & mean per class                            \\
Flowers                  & 102                             & 2,040                              & 6,149                             & mean per class                            \\
STL10                        & 10                             & 5,000                             & 8,000                            & accuracy                                  \\
EuroSAT                         & 10                              & 10,000                             & 5,000                             & accuracy                                  \\
RESISC45            & 45                              & 3,150                              & 25,200                             & accuracy                                  \\
KITTI            & 4                              & 6770                             & 711                             & accuracy                                  \\
Country211            & 211                              & 42,200                             & 21,100                             & accuracy                                  \\
UCF101            & 101                              & 9,537                             & 1,794                            & accuracy                                  \\
Memes            & 2                              & 8,500                             & 500                            & ROC AUC                                  \\
SST2            & 2                              & 7,792                             & 1,821                            & accuracy                                  \\
ImageNet                        & 1000                            & 1,281,167                          & 50,000                            & accuracy                                  \\
\bottomrule
\end{tabular}}

\vspace{-3mm}
\end{table}

%% file: Tables_supp/Data_scaling_linear.tex
\begin{table*}[t]
    \centering
    \small
    \caption{Linear probe on 20 downstream datasets. Pre-training different scale CLIP models on \dsname30M and LAION30M, achieves 1.3\%-3.0\% average performance improvement.}    
    \vspace{-2mm}
    \resizebox{\linewidth}{!}{
    \begin{tabular}{cccccccccccccccccccccccc}
        \toprule
        \belowrulesepcolor{lightgray}
         \rowcolor{lightgray} Model &  Dataset & \rotatebox[origin=lb]{90}{\smash{Food101}} & \rotatebox[origin=lb]{90}{\smash{CIFAR10}} & \rotatebox[origin=lb]{90}{\smash{CIFAR100}} & \rotatebox[origin=lb]{90}{\smash{Birdsnap}} & \rotatebox[origin=lb]{90}{\smash{SUN397}} & \rotatebox[origin=lb]{90}{\smash{Cars}} & \rotatebox[origin=lb]{90}{\smash{Aircraft}} & \rotatebox[origin=lb]{90}{\smash{DTD}} & \rotatebox[origin=lb]{90}{\smash{Pets}} & \rotatebox[origin=lb]{90}{\smash{Caltech}} & \rotatebox[origin=lb]{90}{\smash{Flowers}} & \rotatebox[origin=lb]{90}{\smash{STL10}} & \rotatebox[origin=lb]{90}{\smash{EuroSAT}} & \rotatebox[origin=lb]{90}{\smash{RESISC45}} & \rotatebox[origin=lb]{90}{\smash{KITTI}} & \rotatebox[origin=lb]{90}{\smash{Country}} & \rotatebox[origin=lb]{90}{\smash{UCF101}} & \rotatebox[origin=lb]{90}{\smash{Memes}} & \rotatebox[origin=lb]{90}{\smash{SST2}} & \rotatebox[origin=lb]{90}{\smash{ImageNet}} & \rotatebox[origin=lb]{90}{\smash{\textbf{Average}}} \\
        \aboverulesepcolor{lightgray}
        \midrule
        \multirow{2}{*}{ViT-B/32} 
        & LAION &  76.1 & 94.5 & 80.0 & 47.4 & 70.3 & \colorbox{blue_ours}{82.3} & \colorbox{blue_ours}{45.9} & \colorbox{blue_ours}{74.7} & 80.3 & 89.8 & 89.5 & 95.6 & 95.5 & 84.5 & 72.6 & 15.2 & 76.6 & \colorbox{blue_ours}{56.2} & \colorbox{blue_ours}{60.0} & 64.3 & 72.6 \\
        & \dsname & \colorbox{blue_ours}{81.2} & \colorbox{blue_ours}{95.4} & \colorbox{blue_ours}{81.8} & \colorbox{blue_ours}{48.4} & \colorbox{blue_ours}{74.5} & 73.4 & 45.2 & 74.2 & \colorbox{blue_ours}{84.1} & \colorbox{blue_ours}{91.3} & \colorbox{blue_ours}{90.6} & \colorbox{blue_ours}{97.2} & \colorbox{blue_ours}{96.5} & \colorbox{blue_ours}{89.2} & \colorbox{blue_ours}{74.5} & \colorbox{blue_ours}{19.0} & \colorbox{blue_ours}{82.6} & 55.0 & 56.2 & \colorbox{blue_ours}{68.5} & \colorbox{blue_ours}{73.9} \\
        \midrule
        \multirow{2}{*}{ViT-B/16} 
        & LAION & 82.1 & 95.1 & 81.4 & 57.5 & 73.4 & \colorbox{blue_ours}{87.3} & 47.1 & 76.1 & 84.4 & 91.5 & 92.7 & 96.8 & 95.6 & 86.8 & 70.8 & 17.6 & 80.3 & 59.5 & \colorbox{blue_ours}{65.6} & 68.8 & 75.5\\
        & \dsname & \colorbox{blue_ours}{87.5} & \colorbox{blue_ours}{95.8} & \colorbox{blue_ours}{82.5} & \colorbox{blue_ours}{59.4} & \colorbox{blue_ours}{77.5} & 81.0 & \colorbox{blue_ours}{48.7} & \colorbox{blue_ours}{77.9} & \colorbox{blue_ours}{88.9} & \colorbox{blue_ours}{92.5} & \colorbox{blue_ours}{94.2} & \colorbox{blue_ours}{98.3} & \colorbox{blue_ours}{96.9} & \colorbox{blue_ours}{91.5} & \colorbox{blue_ours}{70.8} & \colorbox{blue_ours}{22.1} & \colorbox{blue_ours}{85.1} & \colorbox{blue_ours}{60.6} & 64.7 & \colorbox{blue_ours}{73.9} & \colorbox{blue_ours}{77.5} \\
        \midrule
        \multirow{2}{*}{ViT-L/14} 
        & LAION & 84.7 & 96.4 & 83.5 & 59.2 & 75.5 & \colorbox{blue_ours}{88.5} & 46.6 & 77.8 & 85.0 & 92.6 & 94.3 & 97.9 & 95.9 & 88.0 & 71.7 & 18.7 & 81.1 & 58.6 & 64.6 & 71.2 & 76.6 \\
        & \dsname & \colorbox{blue_ours}{90.3} & \colorbox{blue_ours}{97.5} & \colorbox{blue_ours}{86.2} & \colorbox{blue_ours}{64.3} & \colorbox{blue_ours}{79.7} & 83.6 & \colorbox{blue_ours}{51.4} & \colorbox{blue_ours}{79.6} & \colorbox{blue_ours}{90.0} & \colorbox{blue_ours}{94.5} & \colorbox{blue_ours}{94.8} & \colorbox{blue_ours}{98.9} & \colorbox{blue_ours}{96.6} & \colorbox{blue_ours}{92.7} & \colorbox{blue_ours}{73.8} & \colorbox{blue_ours}{25.0} & \colorbox{blue_ours}{86.4} & \colorbox{blue_ours}{63.8} & \colorbox{blue_ours}{66.1} & \colorbox{blue_ours}{76.7} & \colorbox{blue_ours}{79.6}\\
        \bottomrule
    \end{tabular}
    }
    \label{table:linear_probe_30M_supp}
\end{table*}

%% file: Tables_supp/Data_scaling_transfer.tex
\begin{table*}[t!]
    \centering
    \small
    \caption{Zero-shot transfer on 20 downstream datasets. Pre-training different scale CLIP models on \dsname30M and LAION30M, achieves 3.5\%-5.5\% average performance improvement.}
    \vspace{-2mm}
    \resizebox{\linewidth}{!}{
    \begin{tabular}{cccccccccccccccccccccccc}
        \toprule
        \belowrulesepcolor{lightgray}
        \rowcolor{lightgray} Model &  Dataset & \rotatebox[origin=lb]{90}{\smash{Food101}} & \rotatebox[origin=lb]{90}{\smash{CIFAR10}} & \rotatebox[origin=lb]{90}{\smash{CIFAR100}} & \rotatebox[origin=lb]{90}{\smash{Birdsnap}} & \rotatebox[origin=lb]{90}{\smash{SUN397}} & \rotatebox[origin=lb]{90}{\smash{Cars}} & \rotatebox[origin=lb]{90}{\smash{Aircraft}} & \rotatebox[origin=lb]{90}{\smash{DTD}} & \rotatebox[origin=lb]{90}{\smash{Pets}} & \rotatebox[origin=lb]{90}{\smash{Caltech}} & \rotatebox[origin=lb]{90}{\smash{Flowers}} & \rotatebox[origin=lb]{90}{\smash{STL10}} & \rotatebox[origin=lb]{90}{\smash{EuroSAT}} & \rotatebox[origin=lb]{90}{\smash{RESISC45}} & \rotatebox[origin=lb]{90}{\smash{KITTI}} & \rotatebox[origin=lb]{90}{\smash{Country}} & \rotatebox[origin=lb]{90}{\smash{UCF101}} & \rotatebox[origin=lb]{90}{\smash{Memes}} & \rotatebox[origin=lb]{90}{\smash{SST2}} & \rotatebox[origin=lb]{90}{\smash{ImageNet}} & \rotatebox[origin=lb]{90}{\smash{\textbf{Average}}} \\
        \aboverulesepcolor{lightgray}
        \midrule
        \multirow{2}{*}{ViT-B/32} 
         & LAION & 58.9 & 85.9 & 63.1 & \colorbox{blue_ours}{17.4} & 54.8 & \colorbox{blue_ours}{61.0} & 4.3 & 36.4 & 65.5 & 82.0 & 41.3 & 91.3 & 40.3 & 43.7 & 24.3 & 7.2 & 47.4 & 51.5 & 50.1 & 44.9 & 48.6 \\
         & \dsname & \colorbox{blue_ours}{67.5} & \colorbox{blue_ours}{89.0} & \colorbox{blue_ours}{65.2} & 15.0 & \colorbox{blue_ours}{60.6} & 39.2 & \colorbox{blue_ours}{7.9} & \colorbox{blue_ours}{37.8} & \colorbox{blue_ours}{70.5} & \colorbox{blue_ours}{84.0} & \colorbox{blue_ours}{42.2} & \colorbox{blue_ours}{93.8} & \colorbox{blue_ours}{59.9} & \colorbox{blue_ours}{61.9} & \colorbox{blue_ours}{27.7} & \colorbox{blue_ours}{10.6}  & \colorbox{blue_ours}{56.7} & \colorbox{blue_ours}{52.5} & \colorbox{blue_ours}{50.1} & \colorbox{blue_ours}{50.9} & \colorbox{blue_ours}{52.1} \\
         \midrule
        \multirow{2}{*}{ViT-B/16} 
         & LAION & 67.6 & 89.1 & 63.5 & \colorbox{blue_ours}{20.8} & 55.7 & \colorbox{blue_ours}{66.9} & 5.4 & 39.0 & 70.2 & 84.9 & 42.9 & 94.3 & 31.1 & 45.4 & \colorbox{blue_ours}{34.0} & 8.7 & 52.2 & 54.5 & 50.6 & 49.4 & 51.3 \\
         & \dsname & \colorbox{blue_ours}{75.8} & \colorbox{blue_ours}{89.6} & \colorbox{blue_ours}{64.7} & 18.9 & \colorbox{blue_ours}{64.3} & 48.2 & \colorbox{blue_ours}{7.9} & \colorbox{blue_ours}{41.2} & \colorbox{blue_ours}{76.0} & \colorbox{blue_ours}{87.5} & \colorbox{blue_ours}{45.2} & \colorbox{blue_ours}{95.1} & \colorbox{blue_ours}{56.8} & \colorbox{blue_ours}{64.3} & 27.1 & \colorbox{blue_ours}{13.1} & \colorbox{blue_ours}{59.1} & \colorbox{blue_ours}{54.5} & \colorbox{blue_ours}{54.0} & \colorbox{blue_ours}{55.9} & \colorbox{blue_ours}{54.9} \\
         \midrule
        \multirow{2}{*}{ViT-L/14} 
         & LAION & 70.8 & 88.8 & 69.5 & \colorbox{blue_ours}{22.8} & 61.6 & \colorbox{blue_ours}{69.7} & 4.9 & 40.8 & 68.0 & 87.3 & 42.2 & 95.3 & 41.5 & 53.7 & 25.9 & 10.4 & 54.7 & 54.1 & 51.8 & 51.5 & 53.3 \\
         & \dsname & \colorbox{blue_ours}{80.7} & \colorbox{blue_ours}{94.1} & \colorbox{blue_ours}{73.1} & 20.9 & \colorbox{blue_ours}{66.4} & 53.6 & \colorbox{blue_ours}{10.1} & \colorbox{blue_ours}{48.1} & \colorbox{blue_ours}{72.8} & \colorbox{blue_ours}{89.4} & \colorbox{blue_ours}{49.8} & \colorbox{blue_ours}{96.2} & \colorbox{blue_ours}{68.5} & \colorbox{blue_ours}{70.1} & \colorbox{blue_ours}{32.2} & \colorbox{blue_ours}{15.3} & \colorbox{blue_ours}{63.9} & \colorbox{blue_ours}{54.1} & \colorbox{blue_ours}{56.9} & \colorbox{blue_ours}{59.5} & \colorbox{blue_ours}{58.8}\\
        \bottomrule
    \end{tabular}
    }
    \label{table:zeroshot_classfication_30M_supp}
\end{table*}

%% file: Tables_supp/Data_scaling_robustness.tex
\begin{table}[t!]
\centering
\caption{Zero-shot robustness comparison. Pre-training different scale CLIP models on \dsname30M and LAION30M, achieves 4.2\%-8.6\% average performance improvement.}
\resizebox{\linewidth}{!}{
    \begin{tabular}{cccccccc}
        \toprule
        \belowrulesepcolor{lightgray}
        \rowcolor{lightgray} Model  & Dataset& IN-V2 & IN-A & IN-R  & ObjectNet & IN-Sketch & \textbf{Average}\\
        \aboverulesepcolor{lightgray}
        \midrule \multirow{2}{*}{ViT-B/32} 
        & LAION & 37.5 & 8.9 & 54.4 & 35.5 & 31.8 & 33.6 \\
        & \dsname & \colorbox{blue_ours}{42.9} & \colorbox{blue_ours}{16.1} & \colorbox{blue_ours}{56.5} & \colorbox{blue_ours}{41.5} & \colorbox{blue_ours}{31.9} & \colorbox{blue_ours}{37.8} \\
        \midrule \multirow{2}{*}{ViT-B/16} 
        & LAION & 42.4 & 12.8 & 60.3 & 40.2 & 34.8 & 38.1 \\
        & \dsname & \colorbox{blue_ours}{48.0} & \colorbox{blue_ours}{24.1} & \colorbox{blue_ours}{63.1} & \colorbox{blue_ours}{46.7} & \colorbox{blue_ours}{36.8} & \colorbox{blue_ours}{43.8} \\
        \midrule \multirow{2}{*}{ViT-L/14} 
        & LAION & 45.1 & 17.1 & 64.9 & 43.1 & 39.0 & 41.8\\
        & \dsname & \colorbox{blue_ours}{52.8} & \colorbox{blue_ours}{34.7} & \colorbox{blue_ours}{71.6} & \colorbox{blue_ours}{50.4} & \colorbox{blue_ours}{42.4} & \colorbox{blue_ours}{50.4} \\
	\bottomrule
    \end{tabular}
    }
\label{tab:robustness_30M_supp}
\end{table}

%% file: Tables_supp/CMP_DreamLIP.tex
\begin{table}[t!]
\centering
\caption{Comparison with DreamLIP's rewriting dataset.}
\resizebox{\linewidth}{!}{
    \begin{tabular}{ccccc}
        \toprule
        \belowrulesepcolor{lightgray}
        \rowcolor{lightgray} Model  & \shortstack{Data Scale}  & \shortstack{Dataset} & \shortstack{Linear probe\\ Avg} & \shortstack{Transfer\\ Avg} \\
        \aboverulesepcolor{lightgray}
        \midrule 
        \multirow{4}{*}{ViT-B/16} & \multirow{2}{*}{15M} 
        & DreamLIP & 73.7 & 43.3  \\
        & & \dsname & \colorbox{blue_ours}{74.9} & \colorbox{blue_ours}{51.6}  \\
        \cmidrule(lr){2-5}
        &  \multirow{2}{*}{30M} & DreamLIP & 77.2 & 53.1  \\
        & & \dsname & \colorbox{blue_ours}{77.5} & \colorbox{blue_ours}{54.9}  \\
	\bottomrule
    \end{tabular}
    }
\label{tab:DreamLIP_cmp}
\end{table}

%% file: Tables/laion_RS_merge.tex
\begin{table}[h!]
\centering
\caption{Data Scaling Comparison. Merged 30M: LAION 15M + \dsname 15M.}
\label{tab: Data_Scaling_Comparison}
\resizebox{\linewidth}{!}{
    \begin{tabular}{lccccc}
        \toprule
        \belowrulesepcolor{lightgray}
        \rowcolor{lightgray} Model  & \shortstack{Dataset}  & \shortstack{Linear probe\\ Avg} & \shortstack{Transfer\\ Avg} & \shortstack{Robustness\\ Avg} \\
         \aboverulesepcolor{lightgray}
        \midrule
        \multirow{6}{*}{ViT-B/16} & \color{dt} LAION15M  & \color{dt} 69.8 & \color{dt} 42.7 & \color{dt} 27.2  \\
         & \color{dt} \dsname15M  & \color{dt} 71.4 & \color{dt} 47.9 & \color{dt} 31.5  \\
         & LAION30M  & 72.6 & 48.6 & 33.6  \\
         & Merged30M  & 73.2 & 48.8 & 34.2 \\
         & \dsname30M  & \colorbox{blue_ours}{73.9} & \colorbox{blue_ours}{52.1} & \colorbox{blue_ours}{37.8} \\
	\bottomrule
    \end{tabular}
    }
\end{table}

%% file: Tables_supp/SimCLR_comparison.tex
\begin{table*}[t!]
    \centering
    \small
    \caption{Linear probe on 12 downstream datasets. Pre-training CLIP~(ResNet50) on image-text pairs achieves 2.1\% performance improvement.}
    \vspace{-2mm}
    \resizebox{0.95\linewidth}{!}{
    \begin{tabular}{cccccccccccccccc}
        \toprule
        \belowrulesepcolor{lightgray}
         \rowcolor{lightgray} Model  & Method & \rotatebox[origin=lb]{90}{\smash{Food101}} & \rotatebox[origin=lb]{90}{\smash{CIFAR10}} & \rotatebox[origin=lb]{90}{\smash{CIFAR100}} & \rotatebox[origin=lb]{90}{\smash{Birdsnap}} & \rotatebox[origin=lb]{90}{\smash{SUN397}} & \rotatebox[origin=lb]{90}{\smash{Cars}} & \rotatebox[origin=lb]{90}{\smash{Aircraft}} & \rotatebox[origin=lb]{90}{\smash{VOC2007}} & \rotatebox[origin=lb]{90}{\smash{DTD}} & \rotatebox[origin=lb]{90}{\smash{Pets}} & \rotatebox[origin=lb]{90}{\smash{Caltech101}} & \rotatebox[origin=lb]{90}{\smash{Flowers}} & \rotatebox[origin=lb]{90}{\smash{\textbf{Average}}} \\
         \aboverulesepcolor{lightgray}
        \midrule
        \multirow{2}{*}{ResNet50} 
        & SimCLR &  68.4 & \colorbox{blue_ours}{90.6} & \colorbox{blue_ours}{71.6} & 37.4 & 58.8 & 50.3 & \colorbox{blue_ours}{50.3} & 80.5 & \colorbox{blue_ours}{74.5} & 83.6 & \colorbox{blue_ours}{90.3} & \colorbox{blue_ours}{91.2} & 70.6 \\
        & Real\&Syn Texts &  \colorbox{blue_ours}{72.5} & 89.1 & 69.0 & \colorbox{blue_ours}{57.1} & \colorbox{blue_ours}{63.6} & \colorbox{blue_ours}{51.4} & 48.1 & \colorbox{blue_ours}{85.5} & 69.7 & \colorbox{blue_ours}{90.5} & 88.1 & 88.8 & \colorbox{blue_ours}{72.7} \\
        \bottomrule
    \end{tabular}
    }
    \label{sim_clr_comparison}
\end{table*}

%% file: Tables_supp/Dataset_comparison.tex
\begin{table*}[t!]
  \centering
  \caption{Current Dataset Comparison. Comparison with large-scale image-text pre-training datasets. }
  \vspace{-2mm}
  \footnotesize  
  \resizebox{0.95\linewidth}{!}{
  \begin{tabular}{l|ccc|cc}
    \toprule
    \textbf{\shortstack{Dataset}} & \textbf{\shortstack{\#Images}}  & \textbf{\shortstack{\#Avg Tokens $/$ Image}} & \textbf{\shortstack{\#Avg Texts $/$ Image}} & \textbf{\shortstack{Text Type}} & \textbf{\shortstack{Source Type}} \\
    \midrule
    CC12M  & \num{12000000} & $-$  &  1 & Realistic  & Website \\
    YFCC15M  & \num{15000000} & 16  & 1 & Realistic & Website \\
    CapsFusion  & \num{120000000} &  $-$  & 1 & Synthetic & Image-Text Pair\\
    LAION400M  & \num{400000000} & 27 &  1 & Realistic & Website\\
    \rowcolor[HTML]{EBEBEB} \dsname 15M & \num{15239498} & 40 &  4 & Realistic \& Synthetic & Interleaved Image-Text \\
    \rowcolor[HTML]{EBEBEB} \dsname 30M & \num{30328852} & 38 & 4 & Realistic \& Synthetic & Interleaved Image-Text \\
    \rowcolor[HTML]{EBEBEB} \dsname 100M & \num{100862786} & 36 & 4 & Realistic \& Synthetic & Interleaved Image-Text \\
    \bottomrule
  \end{tabular}
  }
  
  \label{public_datasets_comparison} 
  \vspace{-2mm}
\end{table*}